\documentclass{article}

\usepackage[final]{neurips_2022}

\usepackage[utf8]{inputenc} 
\usepackage[T1]{fontenc}    
\usepackage{hyperref}       
\usepackage{url}            
\usepackage{booktabs}       
\usepackage{amsfonts}       
\usepackage{nicefrac}       
\usepackage{microtype}      
\usepackage{xcolor}         

\usepackage{algpseudocode,algorithm, algorithmicx}

\usepackage{xspace}
\usepackage{float}
\usepackage{subcaption}
\usepackage{amsmath, bm}
\usepackage{mathtools}
\usepackage{multirow}

\usepackage{wrapfig,lipsum,booktabs}

\newcommand{\system}{\texttt{UnfoldML}\xspace}

\newcommand{\tasktwonoun}{subcategory\xspace}
\newcommand{\tasktwoplural}{subcategories\xspace}
\newcommand{\tasktwo}{\textit{\tasktwonoun-of-interest}\xspace}

\definecolor{mypink2}{RGB}{219, 48, 122}

\definecolor{codegreen}{rgb}{0,0.6,0}
\definecolor{codegray}{rgb}{0.5,0.5,0.5}
\definecolor{codepurple}{rgb}{0.58,0,0.82}
\definecolor{backcolour}{rgb}{0.95,0.95,0.92}
\definecolor{purple}{RGB}{128,0,128}
\definecolor{indigo}{RGB}{75,0,130}
\definecolor{royalblue}{RGB}{65,105,225}
\definecolor{navy}{RGB}{0,0,128}
\definecolor{npurple}{RGB}{199,36,177}
\newif\ifcommenton
\commentonfalse
\ifcommenton
\newcommand{\zc}[1]{\textcolor{red}{[chao: #1]}}
\newcommand{\yx}[1]{\textcolor{blue}{[YX: #1]}}
\newcommand{\alind}[1]{\textcolor{green}{[Alind: #1]}}
\newcommand{\alexey}[1]{\textcolor{indigo}{[AT: #1]}}
\newcommand{\monish}[1]{\textcolor{npurple}{[Monish: #1]}}
\newcommand{\g}[1]{\textcolor{cyan}{[G: #1]}}
\else
\newcommand{\alexey}[1]{}
\newcommand{\zc}[1]{}
\newcommand{\alind}[1]{}
\newcommand{\yx}[1]{}
\newcommand{\monish}[1]{}
\newcommand{\g}[1]{}
\fi

\title{\system: Cost-Aware and Uncertainty-Based Dynamic 2D Prediction for Multi-Stage Classification}

\author{%
  Yanbo~Xu$^{1,*}$, Alind~Khare$^{1,}$\thanks{Authors contributed equally to this research.} , Glenn~Matlin$^{1}$,  Monish~Ramadoss$^{1}$, \\
  \textbf{Rishikesan Kamaleswaran$^{2}$, Chao Zhang$^{1}$, Alexey Tumanov$^{1}$} \\
  $^1$Georgia Institute of Technology, \\
  $^2$ Emory University\\
  Atlanta, GA 
}

\begin{document}

\maketitle

\begin{abstract}
Machine Learning (ML) research has focused on maximizing the accuracy of predictive tasks. 
ML models, however, are increasingly more complex, resource intensive, and costlier to deploy in resource-constrained environments. 
These issues are exacerbated for prediction tasks with sequential classification on progressively transitioned stages with ``happens-before'' relation between them.
We argue that it is possible to ``unfold'' a monolithic single multi-class classifier, 
typically trained for all stages using all data, into a series of single-stage classifiers. 
Each single-stage classifier can be cascaded gradually from cheaper to more expensive binary classifiers that are trained using only the necessary data modalities or features required for that stage. 
\system is a cost-aware and uncertainty-based  dynamic 2D prediction pipeline for multi-stage classification that enables
(1) navigation of the accuracy/cost tradeoff space, 
(2) reducing the spatio-temporal cost of inference by orders of magnitude, and 
(3) early prediction on proceeding stages.
\system achieves orders of magnitude better cost in clinical settings, while detecting multi-stage disease development in real time. 
It achieves within 0.1\% accuracy from the highest-performing multi-class baseline, 
while saving close to 20X on spatio-temporal cost of inference and earlier (3.5hrs) disease onset prediction.
We also show that \system generalizes to  image classification, where it can predict different level of labels (from coarse to fine) given different level of abstractions of a image, saving close to 5X cost with as little as 0.4\% accuracy reduction.
\end{abstract}



\section{Introduction}
\label{sec:intro}
Machine Learning (ML) research has mostly focused on improving prediction accuracy for classification tasks, such as image classification \citep{foret2020sharpness, Xie2017AggregatedRT}, disease risk prediction \citep{feng2020boosting, xu2018raim}, pedestrian detection \citep{Zhang2018OcclusionawareRD, cai2015learning}, etc.
The understandable drive for high accuracy has often resulted in deeper, complex neural networks, which can incur high memory (\textit{spatial} cost) and high latency (\textit{temporal} cost) at inference time.
However, the deployment of ML applications must be cost-aware.
Run time environment like mobile devices or bedside-patient monitors are commonly resource constrained, and applications that can be offloaded to cloud computing always aim for a reduced cloud bill.
In this paper, we focus on developing a pipeline that can balance between high prediction accuracy and low \textit{spatio-temporal} cost in deploying ML classification models.

We consider a scenario of deploying ML classifiers in a multi-stage classification task where one predicted class can progressively transform to a next stage of classes, characterized as ``happens-before'' relationship between classes. This task is commonly observed in many real-world applications. For instance in clinical settings, disease progression is often identified by a series of stage transitions \citep{sperling2011toward, singer2016third}. Detection on early stage of the disease allows doctors to take appropriate actions in time before it enters into late severe stages. In image classifications, recognition from super classes to sub classes in a coarse-to-fine manner has shown improved classification performance \citep{dutt2017improving, lei2017weakly}. In all these applications, general multi-class classifiers~\citep{liu2019data, fagerstrom2019lisep, foret2020sharpness} have been developed by treating all the stages as multi classes and achieved state-of-the-art prediction performance, but at significantly high spatio-temporal cost. Prior work~\citep{wang2017idk} trades off accuracy for low cost but still ignores the key relationship between the classes, so it fails to find the optimal trade-off. 

\begin{figure}
\centering
\begin{subfigure}{.68\textwidth}
  \centering
  \includegraphics[width=\linewidth]{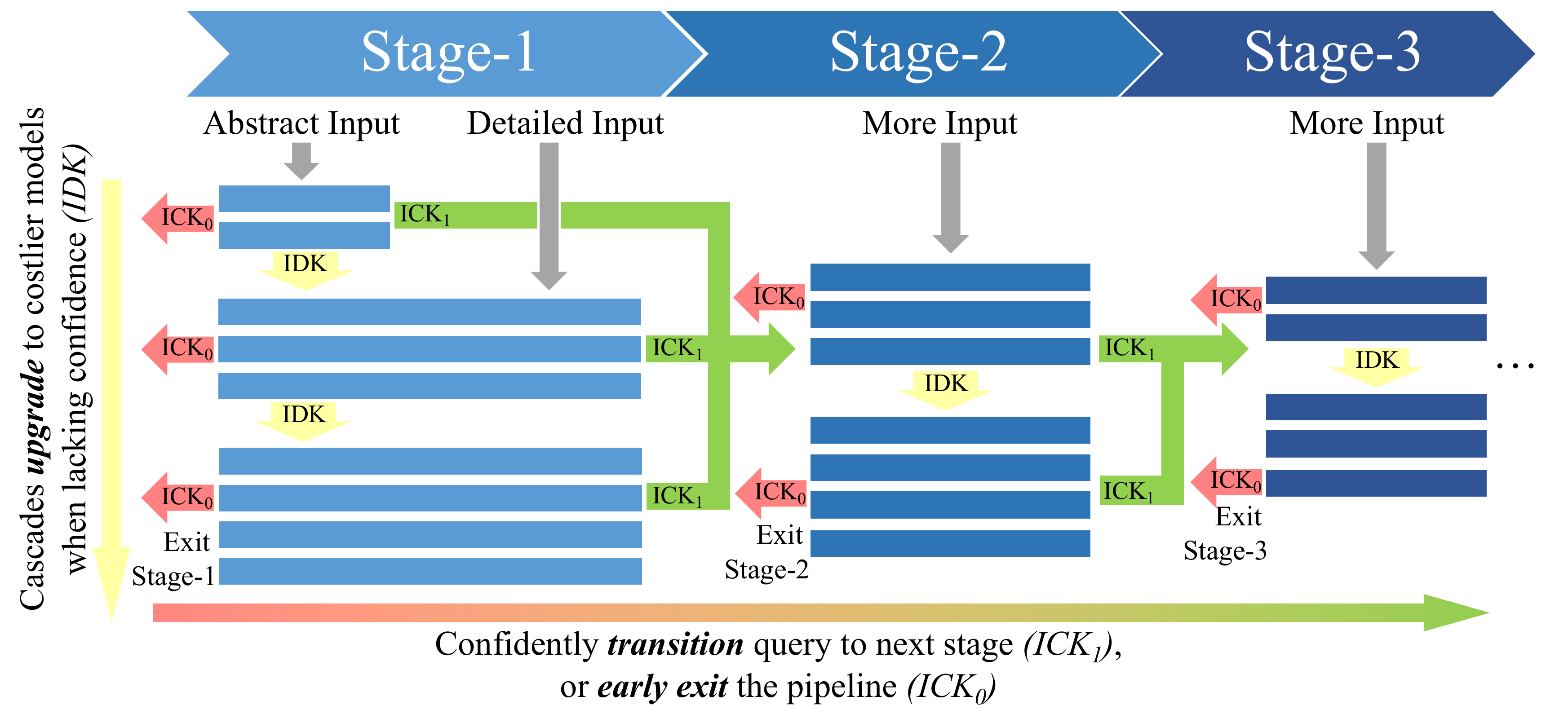}
  \vspace{-1em}
  \caption{\small The 2D query propagation mechanism in \system}
\end{subfigure}%
\quad
\begin{subfigure}{.29\textwidth}
  \centering
  \includegraphics[width=0.9\linewidth]{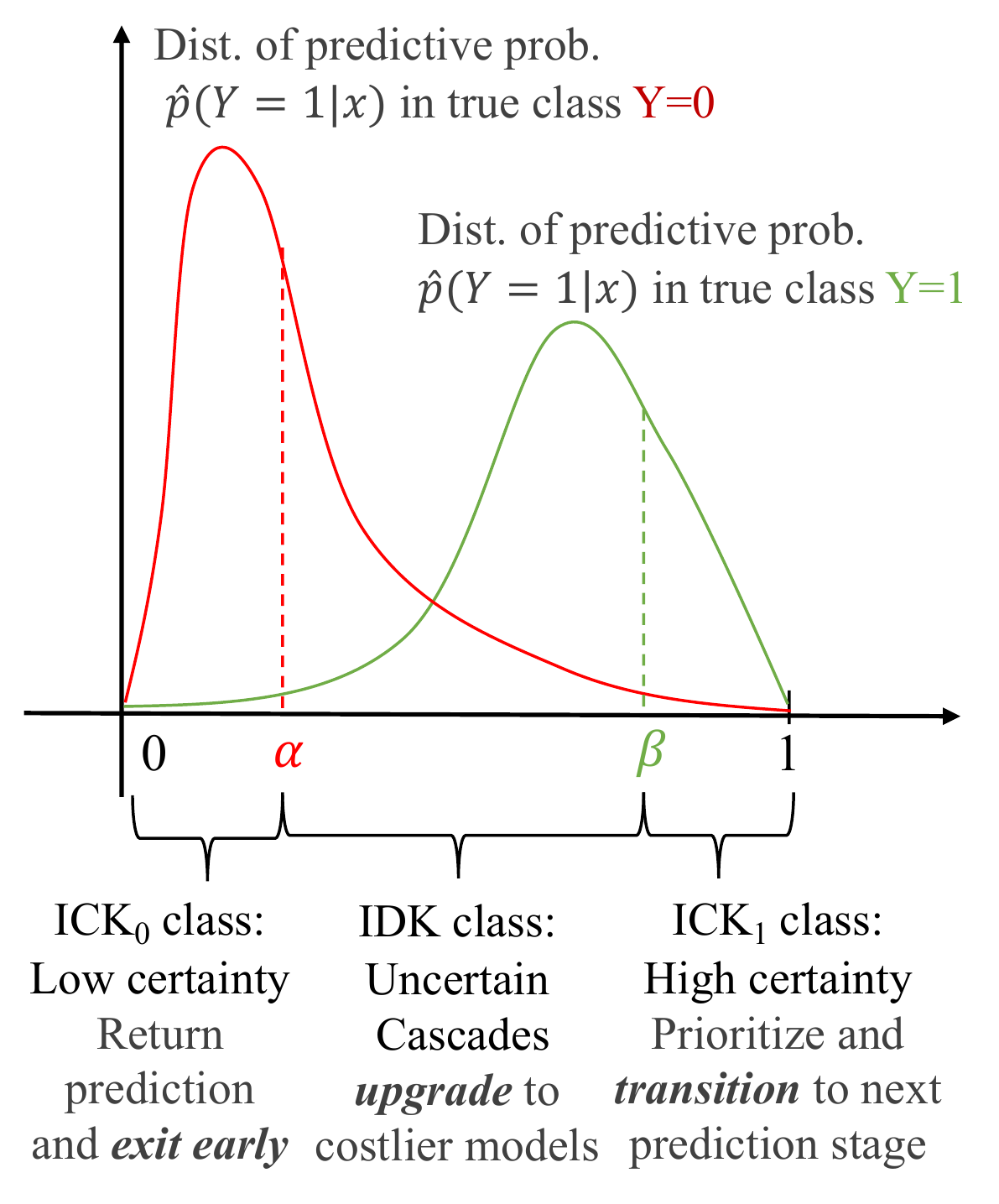}
  \caption{\small $IDK$ and $ICK$ classes}
\end{subfigure}
\caption{\small
    2D uncertainty-based propagation in \system: Queries that are in confidently low risk will return $ICK_0$ and be monitored by cheaper models; queries that are hard to predict will return $IDK$ and be advanced to costlier but more confident models; queries that are in confidently high risk will return $ICK_1$ and be transited to next stage.}
\label{fig:pipeline}
\vspace{-0.1in}
\end{figure}
We propose \system \footnote{Code is available at \url{https://github.com/gatech-sysml/unfoldml}.} : a cost-aware and uncertainty-based prediction pipeline for dynamic multi-stage classification. It ``unfolds'' a monolithic multi-class classifier into a series of single-stage classifiers, reducing its deployment cost. Each single-stage classifier is then cascaded gradually from cheaper to more expensive binary classifiers, further reducing the cost by dynamically selecting an appropriate classifier for an input query. 
Figure~\ref{fig:pipeline} summarizes the two dimensional (2D) query propagation \textit{mechanism} designed in \system: Horizontally it allows a query to transition through multiple stages, and vertically it allows the query to progressively upgrade to costlier models constrained by the pre-specified budget limit. It computes a classifier's prediction confidence on a query then directs the query through one the following three gates: 
1) ``\textit{I confidently know NO}'' ($ICK_0$), which rejects the current query and early exits from the pipeline (exit); 
2) ``\textit{I don't know}'' ($IDK$), which upgrades the query to a higher accuracy but costlier model, producing a more confident result within the budget (vertical cascading); and
3) ``\textit{I confidently know YES}'' ($ICK_1$), which transitions the query to the next stage of the prediction task (horizontal forwarding).  The design of $ICK_0$ gate allows queries to early exit from the \system so it reduces the overall spatio-temporal cost of the pipeline. The $ICK_1$ gate allows queries to faster transition to next stages so it enables early prediction on late stages, which can be critical in clinical settings \citep{reyna2019early}. Overall, the combined 2D propagation mechanism uniquely enables the navigation of the cost/accuracy tradeoff space for searching an optimal set of \textit{policies} for  dynamic model selection at inference time. 

We also propose two training algorithms for learning the optimal \textit{policies} for the designed 2D query propagation in \system. The key idea of the training algorithms is to learn the optimal thresholds on the gating functions defined for $ICK_0$, $IDK$ and $ICK_1$.
The first proposed \textit{hard-gating} algorithm assumes the gating functions to be step functions parameterized by deterministic confidence thresholds. To find the optimal thresholds, it performs bottom-up grid search over a topologically-sorted list of all the models and identifies the thresholds to minimize the prediction loss, while following a cost constraint. 
The secondly proposed \textit{soft-gating} algorithm defines the gating functions to be probabilistic activation functions. It follows a Mixture-of-Expert (MoE) framework to adaptively determine a models' confidence threshold given all the other model's confidence in predicting the same query.
To obviate the cost of running all models in MoE selection, we further propose a Dirichlet Knowledge Distillation (DKD) to run only a cheap multi-label classifier that is trained for distilling the Bayesian predictive uncertainties of all models. 

We summarize the contributions of this paper as follows:
\begin{itemize}
    \item We design a novel 2D query propagation pipeline that ``unfolds'' multi-stage prediction workflows by leveraging the ``happens-before'' relationship between the stages, and achieves a lower-cost prediction pipeline with minimal accuracy degradation.
    \item We propose two learning algorithms to sufficiently navigate the cost/accuracy tradeoff space and search an optimal set of policies for the designed 2D query propagation.
    \item We apply the proposed pipeline to two real-world applications and demonstrate it reduces the spatio-temporal cost of inference by orders of magnitude. 
\end{itemize}

\section{Related Work}
\label{sec:related}

The most relevant work to our proposed method is the one-step $IDK$ cascade
\citep{wang2017idk}, which incorporates prior work of ``\textit{I don't know}'' ($IDK$)
classes \citep{trappenberg2000classification, khani2016unanimous} into cascade
construction and introduce a latency-aware objective into the construction
comparing with previous cascaded prediction frameworks \citep{rowley1998neural,
  viola2004robust, angelova2015real}. Another group of work focus on the problem of feature selection assuming each feature can be acquired for a cost. They train a cascade of classifiers for optimizing the trade-off between the expected classification error and the feature cost. 
  Early solution \citep{raykar2010designing} limits the cascade to a family of linear discriminating functions. \citet{cai2015learning} applies boosting method for cascading a set of weak learners. Recent methods \citep{trapeznikov2013supervised,clertant2019interpretable,janisch2019classification} develop POMDP-based frameworks and incorporate deep Q-learning in training the cascades. In contrast to all of the above work that are only 1-D pipelines for one-step prediction task (can be multi-class classifications), our method extends to a 2D pipeline that can dynamically forward examples to next steps after they are confidently predicted as passed on the current step. Further, we also develop a more efficient pipeline framework based on Mixture-of-Experts (MoE) modeling and knowledge distillation, which can apply gradient decent algorithms for learning the parameters efficiently.

The idea of MoE was originally introduced by \citet{jacobs1991adaptive}, for partitioning the training data and feeding them into separate neural networks during the learning process. This gate decision design is applied into many domains such as language modeling \citep{ma2018modeling}, video captioning \citep{wang2019learning}, multi-tasking learning \citep{ma2018modeling}. It is also used in network architecture searching \citep{eigen2013learning} by setting gate activation on network layers. Sparse gates are introduced in MoE so that it can efficiently select from thousands of sub-networks \citep{shazeer2017outrageously} as well as increases the representation power of large convolutional networks by only using a shallow embedding network to produce the mixture weights \citep{wang2020deep}. We incorporate the idea of sparsely gated MoE \citep{shazeer2017outrageously, wang2020deep} into our prediction framework, and design a soft-gating training algorithm by using ReLU as the sparse gating function and imposing L1-norm regularization on the gating weights for further sparsity.

Confidence criterion has been incorporated into active learning by \cite{li2006confidence} and then extended by \cite{zhu2010confidence}. \cite{lei2014classification} proposed confidence based classifiers that identifies the confident region (like $ICK$ class) and uncertain region (like $IDK$ class) in predictions. Confidence are also introduced into word embedding \citep{vilnis2014word, athiwaratkun2018hierarchical} and graph representations \citep{orbach2012graph, vashishth2019confidence}. Our method posits thresholds on prediction confidence for activating the gates in pipeline expansion.
Bayesian Prior Networks (BPNs) \citep{malinin2018predictive} have been proposed to estimate the uncertainty distribution in model predictions, which is more
computationally efficient than traditional Bayesian approaches
\citep{mackay1992practical,mackay1992bayesian,hinton1993keeping}. We propose Dirichlet Knowledge Distillation (DKD) based on BPNs for distilling prediction uncertainty in large models so that we only need to run a low-cost multi-head model for producing the weights in MoE efficiently.   


\section{Multi-Stage Dynamic Prediction Pipeline}
\label{sec:method}
We introduce a dynamic 2D prediction pipeline \system, which learns the optimal policy for making ``\textit{I confidently know}'' ($ICK$) predictions on sequential multi-stage classification tasks. An optimal policy will effectively trade off prediction accuracy against spatio-temporal costs in order to maximize the overall system accuracy's AUC while staying under user-imposed cost constraints. 

\subsection{Problem Formulation}
\label{sec:formulation}
Given $\bm{x} \in \mathcal{X}$ at time $t$, a multi-stage pipeline decides whether the individual should maintain at the current stage $s$ or progress into the next stage $s+1$ for time $t+1$. If there are a total number of $S$ stages that need to be detected, we train $K$ number of models $m$ for each specific stage $s$ to form a model zoo $\mathcal{M} = \{ \{m_{11},\cdots, m_{1K_1}\}, \cdots, \{m_{S1},..., m_{SK_S}\}\}$. We measure each model's spatio-temporal cost by multiplying the device cost per unit time with the serving time per prediction stage, denoted as $cost(m_{sk})$.

To optimize the limited system resources, we design a 2D \system in the following way: (1) start with the simplest model for prediction of the initial stage on incoming data, and (2) \textbf{\textit{upgrade}} vertically to costlier models on those samples where ``\textit{I don't know}'' ($IDK$), or (3) \textbf{\textit{transition}} horizontally to the next stage of the pipeline samples where ``\textit{I confidently know YES}'' ($ICK_1$) that we have correctly identified the sample at the current stage, otherwise (4) \textbf{\textit{exit}} the pipeline for those ``\textit{I confidently know NO}'' ($ICK_0$) samples that have been identified and discarded. Figure \ref{fig:pipeline} (a) demonstrates the proposed 2D architecture of \system. The central problem of \system is learning the optimal policy for each of the three classes of gating functions with the objective of maximizing high system-wide accuracy while minimizing the prediction cost.

We formulate \system as a decision rule mapping function $m^{casc}: \mathcal{X} \times \mathcal{M} \rightarrow \mathcal{M}$, which takes example $\bm{x}_t$ coming at time $t$ and the current model choice $m_{sk}$ as an input to determine whether the model for the query should take one of the aforementioned three actions: \textit{upgrade} vertically, \textit{transition} horizontally, or \textit{exit} the pipeline.
These decisions rules can be realized by two groups of parameters: a confidence criterion $q: \mathcal{X} \times \mathcal{M} \rightarrow [0,1]$ that measures the confidence score of a model's prediction on a data example, and two gate functions $G^{IDK}$, $G^{ICK_1}: [0,1] \times \{\text{True}, \text{False}\}$ that are applied on to the confidence score per each prediction made by model $m_{sk}$. The two exclusive gates $IDK$ and $ICK_1$ each respectively decides if the current prediction belongs to either an $IDK$ class such that (s.t.) the system will \textit{upgrade} the query to a costlier model while remaining within the user-defined cost budget, or an $ICK_1$ class s.t. the system will \textit{transition} the query to the next stage in the pipeline. The third gate $G^{ICK_0}$ is determined by $\neg G^{IDK} \wedge \neg G^{ICK_1}$. We formalize the decision rule used at each prediction stage as follows
\begin{equation}
\resizebox{.6\hsize}{!}{$
   m^{casc}(\bm{x}_t, m_{sk}; q, G) =
\begin{dcases}
      m_{s(k+1)}, G^{IDK} \wedge \neg  G^{ICK_1}\big(q_{sk}(\bm{x}_t)\big),\\
      m_{(s+1)1}, \neg G^{IDK} \wedge G^{ICK_1}\big(q_{sk}(\bm{x}_t)\big), \\
       m_{sk}, \quad \neg G^{IDK}\wedge \neg  G^{ICK_1}\big(q_{sk}(\bm{x}_t)\big)
\end{dcases}$} \nonumber
\label{eq:casc}  
\end{equation}
where $q_{sk}(\bm{x}_t)$ is a short notation for $q(\bm{x}_t, m_{sk})$ measuring the confidence of model $m_{sk}$'s prediction on data $\bm{x}_t$.
The goal of configuring an optimal pipeline given a restricted computation resource can be formalized as the following optimization problem:
\begin{equation}
    \min_{G} \mathcal{L}\big(m^{\text{casc}}; \mathcal{D}) \quad
    \text{s.t. } cost(m^{\text{casc}}; \mathcal{D}) \leq c,
    \label{eq:all_loss}
\end{equation}
 where $G$ consists of the two gate functions $G^{IDK}$ and $G^{ICK_1}$, $\mathcal{L}$ denotes the end-to-end prediction loss on data $D$ and $c$ is a user-specified cost-constraint for the system. 
\subsection{Gate Parameters Learning}
\label{sec:training}
Given a training data set $\mathcal{D}$, which does not include any data that was used in the training of the models in our model zoo: $\mathcal{D} = \{\big(\bm{x}_i, (y_{i}^1, t_i^1),\cdots, (y_{i}^S, t_i^S)\big)\}_{i=1}^N$, where $\bm{x}_i = (\bm{x}_{i1},\cdots, \bm{x}_{iT_i})$ is an input sequence observed for individual $i$, $y_i^s \in \{0,1\}$ indicates whether the example entered to stage-$s$, and if \textit{yes}, we use $t_i^s \in \emptyset \cup [1, T_i]$ to determine when it entered. We first partition the multi-stage data into $S$ one-stage data sets: $\mathcal{D}^s = \{ (\bm{x}_{i [t_i^{s-1} : t_i^{s}]}, y_i^s == 1) \} \cup  \{ (\bm{x}_{i [t_i^{s-1} : T_i]}, y_i^s == 0) \}$, then divide the learning of $IDK$ and $ICK$ gate parameters into two separable sub-problems: 
\begin{align}
    \textbf{Sub-Objective 1:} & \min_{G_s^{IDK}} \mathcal{L}^{s}\big(m_s^{\text{casc}};\mathcal{D}^s\big) \text{ s.t. } cost(m_s^{\text{casc}}; \mathcal{D}^s) \leq c_s, s=1,\cdots,S\\\nonumber
    \textbf{Sub-Objective 2:} & \min_{G^{ICK_1}} \mathcal{L}\big(m^{\text{casc}};\mathcal{D}, G^{IDK*}\big),
\end{align}
where $\mathcal{L}^s$ is the one-stage prediction loss on data $\mathcal{D}^s$, and $c_s$ is the cost budget that is pre-allocated for stage-$s$ satisfying $\sum_{s}c_s = c$. 

Decomposing the end-to-end optimization problem in Eq. (\ref{eq:all_loss}) into two sub-problems in Eq. (2) allows us to parallelize the training process.
We can efficiently learn each stage's optimal $IDK$ gate parameters by solving Sub-Objective 1, and learn the optimal $ICK_1$ gate parameters by fixing the learnt $IDK$ values in Sub-Objective 2.


\subsubsection{Hard-gating Training Algorithm}
\label{sec:hard_gate}
In this algorithm, we assume $G^{IDK}$ to be a \textit{hard-gating} function that is parameterized by cutoff $\alpha_{sk}$ s.t. the gate is only activated if the level of confidence in the prediction is below a threshold. 
\begin{align}
\setlength\abovedisplayskip{0pt}
 \textit{Hard-gating: } G^{IDK}(q_{sk}(\bm{x}_t)) = \mathbb{I} (q_{sk}(\bm{x}_t) < \alpha_{sk}),
 \label{eq:hard}
 \setlength\belowdisplayskip{0pt}
\end{align}
where $\mathbb{I}(\cdot)$ is an indicator function. Based on the Problem Formulation, a model $m_{sk}$ at stage $s$ can only be activated if $\mathbb{I}(q_{sk}(\bm{x}_t) \geq \alpha_{sk}) \wedge_{j=1}^{k-1} \mathbb{I}(q_{sj}(\bm{x}_t) < \alpha_{sj}) \equiv 1$. Now we write the conditional probability of being at stage-$s$ given input $\bm{x}_t$ as 
\begin{equation}
\resizebox{.85\hsize}{!}{$\text{Pr}(y^s = 1 | \bm{x}_t; m_s^{\text{casc}})  = \sum_{k=1}^{K_s} \mathbb{I}(q_{sk}(\bm{x}_t) \geq \alpha_{sk}) \cdot \prod_{j=1}^{k-1} \mathbb{I}(q_{sj}(\bm{x}_t) < \alpha_{sj}) \cdot m_{sk}(\bm{x}_t),$} \nonumber
\end{equation}
where $m_{sk}(\bm{x}_t) = \text{Pr}\big(y^s = 1 \mid \bm{x_t}; m_{sk}\big).$ The one-stage loss function in Hard-Gating is defined as the negative log-likelihood loss
\begin{equation}
\resizebox{.75\hsize}{!}{$\mathcal{L}_{\text{nll}}^s \big(m_s^{\text{casc}}; \mathcal{D}^s\big) =  - \sum_{i=1}^{N_s} \sum_{t=1}^{T_i} y_{it}^s \cdot \log p_{it}^s  + (1-y_{it}^s) \cdot \log ( 1 - p_{it}^s ),$}\nonumber
\end{equation}
where $p_{it}^s = \text{Pr}(y^s = 1 | \bm{x}_{it}; m_s^{\text{casc}})$, $y_{it}^s = 1$ only if $y_i^s = 1$ and $t \in [t_i^1 - \delta_t, t_i^1]$ for some $\delta_t$ time-steps we wish to early detect the next stage $s+1$.

Algorithm 1 in Appendix \ref{sec:algorithms} describes a bottom-up grid search algorithm for for learning \textit{hard-gating} parameters. We sort the model list $\mathcal{M}^s = \{m_{s1},\cdots m_{sK_s}\}$ in a monotonically increasing order w.r.t AUC and cost which discards any sub-optimal models. For any stage $s$, Algorithm 1 starts by assigning all the $N^s$ samples for that stage to the first model $m_{s1}$, and performing a grid search on the gate parameters $\alpha_{sk}$'s for level $k=1$ to $K_s$. It gradually assigns $IDK$ samples to the next level's model in the list until the cost exceeds the user-defined budget $c_s$ for that stage. In each iteration of searching the cutoff $\alpha_{sk}$, we set an upper bound $maxA$ on the maximum searching value to avoid over-upgrading. Without setting this bound, the algorithm could overfit and put all the samples into the $IDK$ class, then assign them to the next level's model. This consumes the cost quota quickly, and prevents high $IDK$ samples from exploring larger models in the list.


\subsubsection{Soft-gating Training Algorithm}
\label{sec:soft_gate}
In contrast to grid search on the thresholds, we propose a \textit{soft-gating} algorithm formulating an objective function that can be efficiently solved using gradient descent algorithms. In this algorithm, we define the gate function $G^{IDK}$ to be a ReLU function parameterized by a pair of coefficients $(a_{sk}, b_{sk})$ s.t. the gate is only activated if the linear product $a_{sk} \cdot q_{sk}(\bm{x}_t) - b_{sk} > 0$. Formally we define Soft-gating as 
\begin{align}
  \textit{Soft-gating: } G^{IDK}(q_{sk}(\bm{x}_t)) = \text{ReLU}(a_{sk} \cdot q_{sk}(\bm{x}_t) - b_{sk}).
\end{align}
Therefore the conditional probability of being at the stage-$s$ given input $\bm{x}_{t}$ is defined as a mixture of the $K_s$ models available for stage-$s$ prediction:
\begin{equation}
\resizebox{.8\hsize}{!}{
$\text{Pr}(y^s = 1 | x_t; m^{\text{casc}}) = \sum_{k=1}^{K_s} G^{IDK}(q_{sk}(\bm{x}_t)) \cdot m_{sk}(\bm{x}_t) / \sum_{j=1}^{K_s} G^{IDK}(q_{sj}(\bm{x}_t))$.} \nonumber
\end{equation}
Now, the negative log-likelihood loss $\mathcal{L}_{\text{nll}}^s$ becomes solvable using gradient descent algorithms. However, the normalization term in the mixture of experts requires running all the candidates models in the zoo, which conflicts with our cost-saving goal. Therefore, we propose using a Dirichlet Knowledge Distillation (DKD) for training a small surrogate model for each stage to quantify the prediction confidence for each model in the zoo. Then we replace the $q_{sj}(\bm{x}_t)$'s with their estimations $\hat{q}_{sj}(\bm{x}_t)$'s when selecting the expert models to infer with when real predictions are made. The smaller distilled model only needs to be run once per query, requiring much less cost than running all the models. In our experiment, we utilize the first model $m_{s1}$ from each stage, take the embedding of $h_{s1}(\bm{x}_t)$ prior to the last activation layers in $m_{s1}$ and feed it into a 4-layer $K_s$-head Multi-Layer Perceptron (MLP) that is the distilled model. 

The idea of DKD is to posit a Dirichlet prior distribution over the parameters $\bm{\pi}$ characterizing the predicted output categorical distribution (i.e., binomial in our setup) and a surrogate prior network $\bm{f}$ is fit to generate the concentration parameters $\bm{\alpha}_{sk}$ in the prior:
\begin{equation}
\resizebox{.75\hsize}{!}{
$\text{Pr}(\bm{\pi}|\bm{x}_t; m_{sk})   = \text{Dir} (\bm{\pi}|\bm{\alpha}_{sk}); \quad \bm{\alpha}_{sk}  = (\alpha_{sk, 0}, \alpha_{sk,1}) = \bm{f}(\bm{x}_t; m_{sk})$.} \nonumber
\end{equation}
If the learnt concentration parameters yield a flat prior distribution, it means high uncertainty in the model prediction; if they yield a sharp prior distribution, it means low uncertainty. Then an estimation $\hat{q}_{sk}(\bm{x}_t)$ can be computed from the expected predictive probability  
$\hat{p}_{sk} (\bm{x}_{t}) = \mathbb{E}_{\bm{\pi} \sim \text{Dir}(\bm{\pi}|\bm{\alpha}_{sk})} [\pi_1] = \alpha_{sk,1}/(\alpha_{sk, 0} + \alpha_{sk, 1}).$
For training the DKD model, which is a $K_s$-head MLP per each stage in this paper, we define the loss function for head $k$ as a Kullback-Leibler (KL) divergence between the prior distribution and empirical observed distribution: 
\begin{equation}
\resizebox{.55\hsize}{!}{$\mathcal{L}(\bm{\alpha}_{sk}) = \sum_{i=1}^{N_s} \sum_{t=1}^{T_i} \text{KL} \big(\text{Dir}(\bm{\pi}|\bm{\alpha}_{sk}) \mid\mid p_{sk}(\bm{x}_{it})\big),$} \nonumber
\end{equation}
where $p_{sk}(\bm{x}_{it})$ are the real predicted probabilities produced from model $m_{sk}$ on input $\bm{x}_{it}$. Additionally, we also add the cross-entropy loss as an auxiliary loss when training the DKD. 
Once the distilled model is trained, we replace the $q_{sj} (\bm{x}_t)$'s with the estimated values from the model. Then for cost constraining, we ensure the selected models do not exceed user-imposed cost constraint by enforcing sparse gating weights over the model choices. Therefore, we reformulate the objective function in Hard-Gating Sub-Objective 1 as the following Lagrangian function $$\min_{G^{IDK}} \mathcal{L}_{\text{nll}}^s (m^{\text{casc}}; \mathcal{D}_s)  + \lambda \mathcal{L}_{\text{cost}}^s +  \mu \mathcal{L}_{\text{sparse}}^s,$$
where the second term $\mathcal{L}_{\text{cost}}^s$ takes the cost constraint, controlled by $\lambda >0$ and the third term $\mathcal{L}_{\text{sparse}}^s$ imposes further sparse regularization on the gating weights w.r.t their L$_1$ norms, controlled by $\mu > 0$. Formally, we write the last two terms as
$\mathcal{L}_{\text{cost}}^s  = \big(\max(0, cost(m_s^{\text{casc}}) - c_s)\big)^2$ and $\mathcal{L}_{\text{sparse}}^s = \sum_{i,t} \|G^{IDK}(\hat{q}_{sk}(\bm{x}_{it}))\|_1$.
Now we use stochastic gradient descent algorithms to learn the optimal $\bm{a}^s$ and $\bm{b}^s$ minimizing the above loss. Algorithm 2 in Appendix \ref{sec:algorithms} summarizes the \textit{soft-gating} training algorithm.

\subsubsection{Overall Training Algorithm}
To complete the overall training algorithm, we need to learn the optimal gate parameter $G^{ICK_1}$ in Sub-Objective 2. Given the model $m_{sk}$ picked by one-stage $IDK$ cascade (either using \textit{hard-gating} or \textit{soft-gating} algorithms) for a data example $\bm{x}_t$ at stage $s$, we grid search for the optimal thresholds $\theta_{sk}$'s on the predictive probabilities s.t. the type I error on class $ICK_1$ and type II error on $ICK_0$ are both minimized. Several existing methods \citep{liu2012classification,perkins2006inconsistency,unal2017defining,miller1982maximally} have been proposed for minimizing both type I error and type II error in various ways, we pick the \textit{Closet-to-(0,1)} \citep{perkins2006inconsistency} method that finds the optimal threshold achieving the most left upper corner in the ROC curve. Finally, Algorithm 3 in Appendix \ref{sec:algorithms} gives the end-to-end training Algorithm for \system, where we can use either Algorithm 1 or 2 to learn the optimal gating policy.
\section{Experiments}
\label{sec:exp:setup}
We evaluate \system on two real-world tasks. The first task is to predict if and when a patient who was newly admitted into the Intensive Care Unit (ICU) of a hospital will develop sepsis (Stage-1), which can then progress into septic shock (Stage-2). 
The second task is detection of the \tasktwo in a label hierarchy that filters out queries that are not of interest in Stage-1 and then refines the predictions into fine classes in Stage-2.

\subsection{Task 1: Sepsis-Septic Shock prediction}
We use MIMIC-III Critical Care Database \citep{johnson2016mimic}.
The database consists of deidentified health records from over $50,000$ critically ill patients who stayed in the ICUs of the Beth Israel Deaconess Medical Center between 2001 and 2012. 
We detail our data preparation in Appendix \ref{sec:data}
The final cohort includes a total of $34,475$ ICU patients, from which $2,370$ $(6.8\%)$ presented with Sepsis, from which a total of $229$ $(9.7\%)$ progressed into Septic Shock. We randomly split our cohort of patient data into a training set ($70\%$), validation set ($20\%$) and test set ($10\%$). First, we use the training set to train a set of models to formulate a model zoo. Next, we use the validation set to train the \system policy. Finally, we use the test set to evaluate performance.
 
\subsubsection{Experimental Setup}
\noindent \textbf{Model Zoo.} We construct our model zoo by training two sets of binary classifiers for Stage-1 (Sepsis) and for Stage-2 (Septic Shock) using a variety of architectures and features. We select CPU-based models such as Logistic Regression, Decision Tree and Random Forest, and GPU-based models such as LSTM (Long Short-Term Memory) that is thus far the state-of-art approach in the early detection of Sepsis and Septic Shock \citep{fagerstrom2019lisep, liu2019data}. For LSTM, we vary the hidden size ranging from 100 to 400, the number of layers from 1 to 4. We also vary the window of patient data we input over 1, 6, and 12 hours.
For each stage and model architecture we train models with different combinations of feature sets based on the collection modality. All models start with with basic demographic features and vital signs, and are then extended with additional features including results from lab tests, beside monitoring, and medication/IV treatments. 

\noindent
\textbf{Confidence Measures}
We consider four choices for measuring the confidence $q$ of a model prediction. Given $p$, the model's predictive probability of output $Y$ being $1$, we define
\begin{itemize}
    \item  Max probability: $\max\big(1-p, p\big)$,
    \item Entropy: $\big(p\cdot \log p + (1-p) \cdot \log(1-p)\big)$,
    \item Entropy of expected: $- \frac{\alpha_0 }{\alpha_0 + \alpha_1} \cdot \big(\psi(\alpha_0) - \psi(\alpha_1 )\big) + \psi(\alpha_0 + \alpha_1)$,
    \item Mutual Information: $ \text{Entropy } - \text{Entropy of expected}$, 
\end{itemize}
\noindent \textbf{Spatio-Temporal Cost.}
Given the trained model zoo, we profile spatio-temporal costs for each of the models. The spatio-temporal cost is the total time spent in each hardware due to inference/forward-pass calls of the models (temporal cost) multiplied with the hardware's cost per unit time (spatial cost). This cost serves as a proxy for the real dollar cost as it is the basis for pricing models in cloud offerings such as AWS On-Demand, Lambda, or Spot. 

\noindent \textbf{Baseline.}
To our knowledge, \system is the first system to provide 2-dimensional cascading predictions. A reasonable baseline for our method are models that frame the multi-stage sequential task as a multi-class classification task by ignoring the \textit{happen-before} relationship between the stages. The baseline's goal is to classifying patients into one of the three classes: Non-Septic, Septic, and Septic Shock. We use a state-of-the-art LSTM which is widely used in prediction of clinical time-series. We evaluate performance LSTM's across a variety of spatio-temporal costs by changing the number of layers and hidden sizes in the architecture.

$\bullet$ \textit{Multi-class LSTM}: One unified multi-class model works end to end for predicting the multi classes. Prior work \citep{wang2017idk} implements a 1-D prediction cascade on \textit{IDK} classes. It uses a similar \textit{hard-gating} algorithm to learn hard thresholds on prediction entropy for making decisions if the system should cascade to costlier models. So we reduce \system to a 1-D pipeline by removing its horizontal \textit{ICK} transition and compare the \textit{soft-gating} algorithm with this baseline. We evaluate them on the two single-stage classification tasks on sepsis and septic shock predictions respectively.

$\bullet$  \textit{Single-Stage binary classifiers}: CPU-based models such as Logistic Regression, Decision Tree, Random Forest and TREWScore \citep{henry2015targeted}, a cox proportional hazards model that is well-known in early detection of septic shock; GPU-based models such as LSTMs.

$\bullet$ \textit{IDK-cascade} \citep{wang2017idk}: Prior work of 1-D prediction cascade on \textit{IDK} classes using a hard-gating like algorithm.

\subsubsection{Evaluation and Results}
\noindent\textbf{End-to-End Classification Performance.}
To evaluate the prediction performance on our multi-stage task, we treat \system as a multi-class classifier which predicts one of the  same three classes as defined in Baseline. \system generates predictions at time-step for every patient, we take the maximum of the predictive probabilities along the prediction horizon for each patient and normalize them with a sum of 1. We compute the multi-class ROC AUC scores by averaging the pairwise ROC AUCs (known as one-vs-one) of each classes. 
\label{sec:exp:result}

\begin{figure}[t]
\centering
\begin{subfigure}{.45\textwidth}
  \centering
  \includegraphics[width=0.95\linewidth]{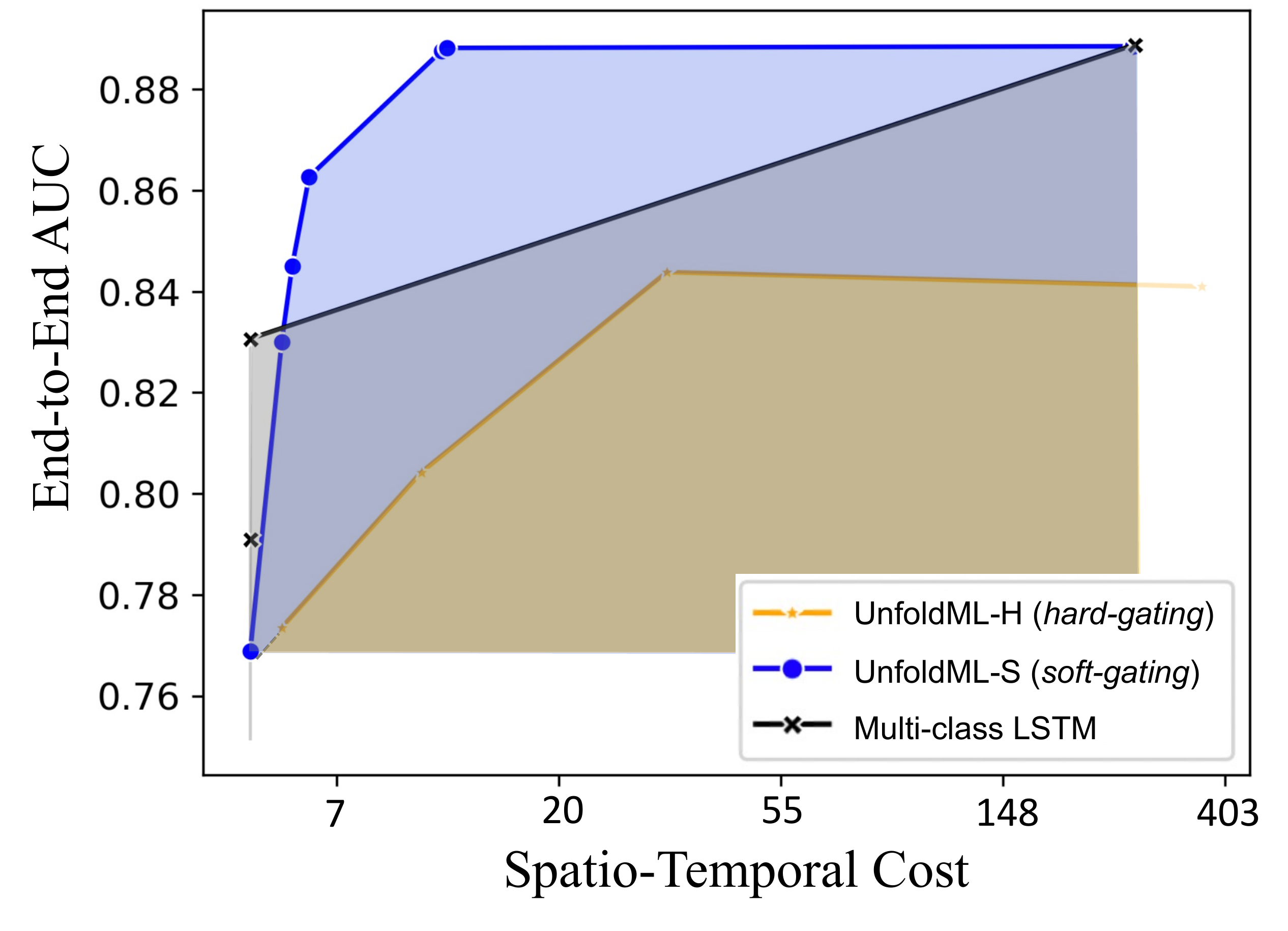}
  \caption{\small Convex hull comparison for the set of points searched by different methods in the  End-to-End AUC vs. Spatio-Temporal Cost tradeoff space.}
  \label{fig:s_search}
\end{subfigure}%
\quad
\begin{subfigure}{.45\textwidth}
  \centering
  \includegraphics[width=0.95\linewidth]{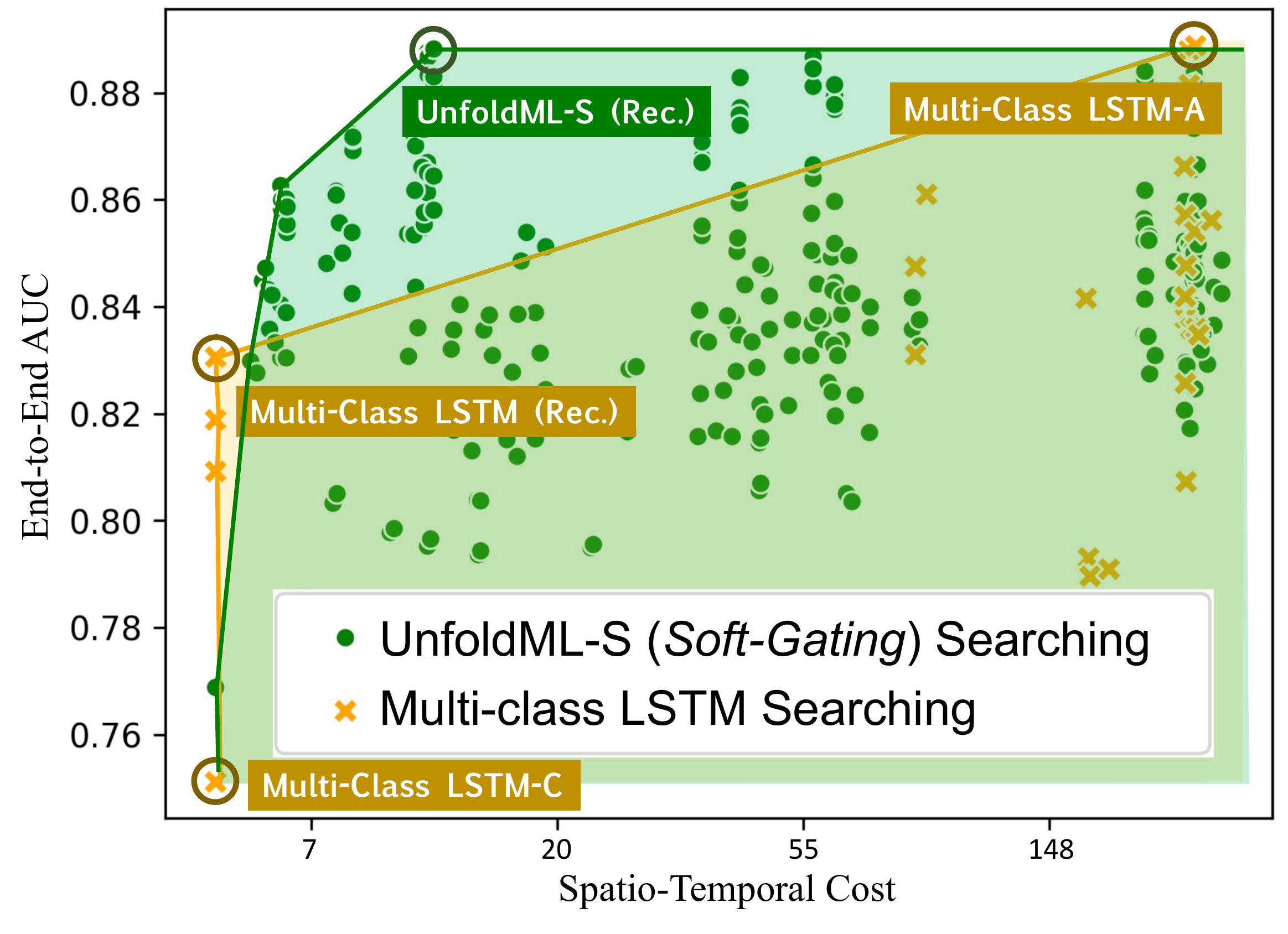}
  \caption{\small Search the trade off between End-to-End AUC and Spatio-Temporal cost by varying the cost constraint (x-axis is in a logarithmic scale).}
  \label{fig:rec}
\end{subfigure}
\caption{\small Tradeoff space of the End-to-End AUC vs. Spatio-Temporal Cost.}
\label{fig:test}
\end{figure}

\noindent\textbf{Better Cost-AUC tradeoff.} 
We evaluate the model performance in trading off between the end-to-end AUC and spatio-temporal cost as follows: we vary the user-defined cost constraint $c$ in Eq. \ref{eq:all_loss}, and train \system repeatedly while searching the trade off in a 2D space. Each run contributes a point in the scatter plot of Figure \ref{fig:rec}. We compute the convex hull of the searched points in the set and demonstrate the resulting area of the hull in Figure \ref{fig:s_search}. We train our baseline method repeatedly by varying the LSTM architecture and plot in the same way as in Figure \ref{fig:s_search}. In comparison, \system-S (\textit{soft-gating}) searches significantly higher AUC regions than the multi-class LSTM baseline and performs more consistently than the \textit{hard-gating} algorithm.
To quantify the trade-off evaluation in Figure \ref{fig:s_search}, we calculate the area of the convex hull searched by each method (Table \ref{tab:hull}). 

\begin{table}
\centering
\resizebox{0.75\linewidth}{!}{%
\begin{tabular}{l|c|c|c}
\toprule
& \system-H & \textbf{\system-S} &  Multi-class LSTM\\
\midrule
Approx. area of the convex hull & 17.58 & \textbf{31.17} & 24.20\\
 \bottomrule
\end{tabular}}
\caption{\small
    The approximated area of the convex hull searched by different methods in Figure \ref{fig:s_search}.}
\label{tab:hull}
\end{table}%
\system-S presents the highest score comparing to the baseline with \system-H. In addition, we pick several critical points from the searched convex hull in Figure \ref{fig:rec} and present them in Table \ref{tab:e2e}. The recommended configuration for systems that effectively trade off spatio-temporal cost and end-to-end AUC is to select models along the frontier line drawn through the points that form the top-left corner of the hull.
As shown in Table \ref{tab:e2e}, an optimal configuration for our pipeline \system-S (Rec.) can achieve comparable end-to-end AUC score ($88.9\%$) with the most accurate baseline Multi-Class LSTM-A ($88.8\%$), while operating at a 19.6X spatio-temporal cost reduction. In comparison with \system-A, which naively uses the most accurate single-stage models in each stage, \system-S outperforms the end-to-end AUC while providing a 22.7X reduction in spatio-temporal cost. Alternatively, \system-S (Rec.) outperforms Multi-class LSTM (Rec.) at an AUC gain of $5.3\%$ with only 1.3X more cost.
The improvement in end-to-end AUC scores from \system is a result of the dynamic nature of inference pipeline, which selects the optimal pathway for each patient. The savings in spatio-temporal cost can be attributed to the usage of low cost models when they are confident enough, in contrast, baselines (\system-A, Multi-Class LSTM-A) always use the most accurate model.

\begin{table}[t]
\centering
   \resizebox{0.75\linewidth}{!}{%
\begin{tabular}{l|cc|c}
\toprule
\multicolumn{1}{c|}{} & \multicolumn{3}{c}{\textbf{Sepsis-Septic Shock Prediction}} \\
 \multirow{2}{*}{}  & end-to-end  & Spatio-Temporal Cost Per & Early Prediction on\\ 
 & AUC (\%) & Inference Call (\$) & Septic Shock (hr)\\
\toprule
Multi-class LSTM-C  & 75.1 & \textbf{5.1}  & 12.6\\
Multi-class LSTM-A & \textbf{88.9} & 269.0  &  24.0 \\
Multi-class LSTM (Rec.) & 83.5 & 6.0  &  15.4 \\
\midrule
 \system-C
 & 76.9 & 5.8 & 22.9\\
\system-A
 & 87.2 & 311.8 & 16.4\\
\system-H (Rec.) & 84.4 & 34.2  & 14.2\\
\textbf{\system-S (Rec.)} & 88.8 & 13.7  & \textbf{26.1}\\
\bottomrule
\end{tabular}}
\caption{\small
    End-to-end performance comparison on two-stage prediction (`-C' denotes the model choice of \textit{the cheapest and least accurate}; `-A' denotes the model choice of \textit{the costliest and most accurate}; `-Rec.' denotes the recommended model choice with a good \textit{trade-off between cost and accuracy} in Figure \ref{fig:rec}; `-S' denotes \textit{soft-gating}; `-H' denotes \textit{hard-gating}.}
\label{tab:e2e}
\end{table}%
\noindent \textbf{Single-Stage Performance.} We report single-stage performance on Sepsis and Septic Shock predictions in Table \ref{tab:stage}. For both predictions, \system achieves a significant  reduction in costs with only a marginal loss of AUC compared to the highest accuracy baseline LSTM-A --- 32.3x lower spatio-temporal cost with 1.7\% lower AUC for Sepsis prediction and 26.8x lower costs with 0.7\% lower AUC for Septic Shock prediction. The baseline IDK-Cascade can achieve comparable AUC as \system-S, but requires 3.4x higher costs.

\begin{table}[t]
\centering
\resizebox{.9\linewidth}{!}{%
\begin{tabular}{l|cc|cc}
\toprule
\multicolumn{1}{c|}{} & \multicolumn{2}{c|}{\textbf{Sepsis Prediction}} & \multicolumn{2}{c}{\textbf{Septic Shock Prediction}} \\
   & Single-Stage & Spatio-Temporal Cost & Single-Stage & Spatio-Temporal Cost \\ 
   & AUC (\%) &  Per Inference Call (\$) & AUC (\%) &  Per Inference Call (\$) \\ 
\toprule
TREWScore \citep{henry2015targeted} & - & - & 83.0 & \textbf{2.3}\\
Logistic Regression & 74.5 & \textbf{2.3} & 87.0 & \textbf{2.3} \\
Decision Tree & 70.4 & 2.4 & 83.1 & 2.4  \\
Random Forest & 75.7 & 148 & 85.2 & 148\\
\midrule
LSTM-C & 88.6 & {5.1} & 90.3 & 5.1 \\
LSTM-A & \textbf{92.8} & 268  &  \textbf{96.9} & 268\\
\midrule
IDK-Cascade \citep{wang2017idk} & 91.7 & 28.1 & 95.2 & 34.0\\
\textbf{\system-S} & \textit{91.1} & \textit{8.3} & \textit{96.2} & \textit{10.0}\\
\bottomrule
\end{tabular}}
\caption{\small Single-stage performance}
\label{tab:stage}
\end{table}
\noindent\textbf{Better Early-hour Prediction.} 
Table 2 also reports the average early prediction, showing
\system-S predicts septic shock earlier than all other methods. It predicts 2.1 hrs prior to the strongest baseline Mutli-class LSTM-A. It benefits from the multi-stage cascaded prediction in \system such that the system can exit early from the sepsis stage once it is in the $ICK_1$ class and proceed to shock prediction before sepsis is truly diagnosed. 



\subsection{Task 2: Subcategory Classification} 
To demonstrate the generalizability of \system we conduct experiments on a computer vision task where we wish to detect a \tasktwo. In the \tasktwo task, we wish to accurately predict if an image falls within a specific \tasktwonoun of a dataset with classes that form a hierarchical structure, while still ensuring our spatio-temporal costs remain within the user-defined budget. In this task, each query is an image from which \system is asked to predict if the image belongs to a \tasktwonoun in the label hierarchy we are interested in.
For our experiment using the CIFAR-100 dataset \citep{krizhevsky2009learning}, we define two separate \tasktwo tasks based on the real-world system applications of computer vision systems. We want to identify images of a specific \tasktwonoun from one of two chosen categories: `people' (baby, boy, girl, man, woman) and `vehicles' (bicycle, bus, motorcycle, pickup truck, train). CIFAR-100 consists of 60,000 32x32 colour images in total of 100 `fine' classes. There are 500 training images and 100 testing images per class. In addition, each image also is assigned a `coarse' label indicating which category (such as people, vehicles, trees, etc) it belongs to. In our experiment, we do a multi-class classification on the 'coarse-granularity' labels in Stage-1 to detect the category, then we do multi-class classification on the original `fine-granularity' CIFAR-100 labels in the second step to identify the \tasktwo.

\subsubsection{Experimental Setup}
Our model zoo is made of WideResNets (WRN) using Sharpness-Aware Minimization \citep{foret2020sharpness} which is cited as having state-of-the-art performance on the CIFAR-100 image recognition task. For each stage we train models with widths from $[2,4,6,8,10]$ and depths from $[16, 22, 24, 28]$.
For Stage-1 we train $K_{1}$ number of multi-class classification models using the full dataset of CIFAR-100 with coarse-granularity labels as our target \tasktwoplural, resulting in a total of 20 classes. We train models for Stage-1 using the full 32x32 image and as well as random crops of size 24x24 or 16x16. For Stage-2 we train $K_{2}$ multi-class models to identify the next \tasktwonoun that \textit{happens-after} the coarse \tasktwonoun from Stage-1. We train our Stage-2 models only using data with the coarse-granularity label for our \tasktwonoun in Stage-1. We train Stage-2 models using the full 32x32 image or random crop of 24x24. We profile each model by computing the number of Multiply-Accumulate Operations (MACs) and use it as proxy measure of our spatio-temporal cost.

\subsubsection{Evaluation and Results}
We compare \system against a multi-class WRN classifier baseline which directly predicts labels at the fine-granularity. We train the baseline with a width factor of 10 and a depth of 28 using the full-sized 32x32 image. We evaluate both accuracy and MACs, and report the results in Table \ref{tab:cifar}. The results show that \system is able to find the optimal gating policy for the pipeline and achieve a spatio-temporal cost-savings of 6.9X with 1\% reduction in accuracy for `people' \tasktwonoun classification; we achieve a cost-savings of 4.7X with 0.4\% reduction in accuracy for `vehicle' \tasktwo identification. Thus, \system achieves a savings in spatio-temporal cost in for this task without compromising the overall accuracy. 
\system can accomplish this because when it is confident that queries do not belong to our \tasktwo ($ICK_0$), we are able to \textit{early exit} the query from the prediction pipeline. When \system is not confident ($IDK$), it \textit{upgrades} queries to costlier and more accurate models. Only when \system can confidently predict the \tasktwonoun ($ICK_1$) in Stage-1 will it \textit{transition} the query to Stage-2 and make the final prediction. This results in a significant cost-savings without compromising accuracy. 
\begin{table}[t]
\centering
\resizebox{0.65\linewidth}{!}{%
\begin{tabular}{l|cc|cc}
\toprule
 & \multicolumn{2}{c|}{Subcategory `People'} & \multicolumn{2}{c}{Subcategory `Vehicles'} \\
\toprule
  & Accuracy (\%) & Macs (10M) & Accuracy (\%)  & Macs (10M)\\ 
WRN-A & \textbf{98.1} & 525.0  & \textbf{99.3}  & 667.0 \\
\textbf{\system-S} & 97.1 & \textbf{77.0}  & 98.9 & \textbf{142.3} \\
\bottomrule
\end{tabular}}
\caption{\small
    Performance on subcategory classification}
\label{tab:cifar}
\end{table}

A recent alternative method for trading-off accuracy and cost in image classification is CwCF (Classification with Costly Features) \citep{janisch2019classification}, which traverses the trade-off space by varying a cost parameter that limits the number of selected features and uses Deep Q-Learning (DQL) model to train a feature-cost-aware classifier. In addition, it also includes a pre-trained cost-unaware High-Performance Classifier (HPC), which is called when it decides to include all the features.
In order to establish a fair baseline against \system, we used the most accurate multi-class WRN for the HPC in CwCF.
We vary their cost parameter lambda for searching the optimal trade-off point in the accuracy-cost space, however, it only returns two extreme data points that gives either low accuracy of 1\% with only $<10$ features or high accuracy as the WRN-A shows in Table \ref{tab:cifar} with all the features. We observe similar failure in traversing the trade-off space for CIFAR-10 in Figure 4(c) of their paper, so we do not include this as a baseline for this task.

\section{Conclusion}
\label{sec:conclusion}
ML models, including for healthcare applications, are growing exponentially in size and cost of inference. 
This is problematic for resource constrained hospital environments. Costlier, monolithic models require expensive hardware, and can not fit on bed-side compute or even on site compute clusters with clinical implications.
~\system proposes a set of mechanisms and policies to address the growing cost of monolithic classifiers for healthcare applications. It implements a query propagation mechanism that ``unfolds'' a monolithic multi-class classifier into a sequence of single-class classifiers, each with its own cascade progressively more complex models. 
Each query is allowed to  
(1) confidently exit the pipeline with an $ICK_0$, 
(2) transition horizontally to the next stage in the pipeline with an $ICK_1$, or 
(3) upgrade vertically to a more complex model within the horizontal stage with an $IDK$. This mechanism is coupled with a set of policies, such as soft-gating, that set the thresholds at which the state transitions occur for queries to the system.
\system builds on a fundamental insight that classes may have a ``happens before'' relationship between them, which can be leveraged to ``unfold'' a classifier, leading to savings in spatio-temporal cost (how much resource used for how long) and clinically significant earlier onset prediction. \system improves the frontier of optimality in the cost/accuracy tradeoff space and is able to nearly match (within 0.1\%) SoTA AUC performance for septic shock prediction at the $\frac{1}{20}^{th}$ of the baseline cost. \system and the application of the ``happens before'' insight generalizes to computer vision tasks with 5x cost savings gained for a mere 0.4\% drop in accuracy.
Limitations of this work include demonstrations on more than 2 stages tasks.

\section*{Acknowledgements}

This material is based upon work supported by the National Science Foundation under Grant Numbers  NSF IIS-2106961, CAREER IIS-2144338, and CCF-2029004. We would also like to acknowledge Dr. Kevin Maher and Dr. Alaa Aljiffry of Children's Healthcare of Atlanta for their medical insights and clinical guidance as well as the Neurips'22 Area Chairs and reviewers for their insightful feedback, which contributed to the improved quality of this paper.
{\bf Disclaimer: } Any opinions, findings, and conclusions or recommendations expressed in this material are those of the authors and do not necessarily reflect the views of the National Science Foundation.




\bibliographystyle{ACM-Reference-Format}
\bibliography{reference}
\clearpage
\section*{Appendix}
\subsection{Training Algorithms}
\label{sec:algorithms}
\textbf{Algorithm} 1 presents the Hard-Gating Training Algorithm, \textbf{Algorithm} 2 presents the Soft-Gating Training Algorithm, and \textbf{Algorithm} 3 summarizes the End-to-End Training Algorithm.

\begin{algorithm}[]
 \caption{\textit{Hard-gating} Algorithm for In-Stage $IDK$ Cascade}
\begin{algorithmic}[1]
\Statex \textbf{Input}
\Statex \hspace*{\algorithmicindent} $\mathcal{D}^s$: Training data containing $N^s$ samples in stage-$s$
\Statex \hspace*{\algorithmicindent} $\mathcal{M}^s$: Sorted list of the models trained for stage-$s$
\Statex \hspace*{\algorithmicindent} $\mathcal{C}$: Dictionary of models' spatio-temporal costs
\Statex \hspace*{\algorithmicindent} $c_s$: User-defined budget of spatio-temporal cost for stage-$s$
\Statex \hspace*{\algorithmicindent} $q$: Confidence function
\Statex \hspace*{\algorithmicindent} $maxA$: Value for the upper bound of the cutoffs to avoid over-fitting
\Statex \hspace*{\algorithmicindent} $nBins$: Number of bins for the grid search
\Statex \textbf{Output}
\Statex \hspace*{\algorithmicindent} $\bm{\alpha}^*_s$: The optimal IDK cutoff vector for stage-$s$
\Procedure{HardGating}{$\mathcal{D}^s$, $\mathcal{M}^s$, $c_s$, $\mathcal{C}$, $q$, $maxA$,  $nBins$}

\State $\bm{\alpha}^*_s = []$, $ModelAssign = \bm{1}$, $cost = \sum_{i,t} \mathcal{C}[m_{s1}]$
\If{$cost > c_s$} \Return $\bm{\alpha}^*_s$
\EndIf
\For{$k$ in range$(K_s-1)$} \Comment{Bottom-up search}
\State $Idx4k \leftarrow \cup I( ModelAssign[i,t] == k)$.
\If{$Idx4k$ is $\emptyset$} \textbf{break}
\EndIf
\State $minQ \gets \min_{Idx4k} \big\{q_{sk}(\bm{x}_{it})\big\}$
\State $maxQ \gets \min(maxA, \max_{Idx4k}
\big\{q_{sk}(\bm{x}_{it})\big\})$.
\State $\alpha^*_{sk} \gets minQ$.
\For{$\alpha_{sk}$ in $LinSpace(minQ, maxQ, nBins)$}
    \State $IDK \gets  \cup_{Idx4k} I\big(q_{sk}(\bm{x}_{it}) \in [\alpha^*_{sk}, \alpha_{sk})\big)$
    \If{$IDK$ is not $\emptyset$}
        \If{$cost + \sum_{IDK} \mathcal{C}[m_{sk+1}] - \mathcal{C}[m_{k}] > c_s$}
        \State $\hat{\bm{\alpha}}_s \gets \bm{\alpha}^*_s + [\alpha^*_{sk}]$; \Return $\bm{\alpha}^*_s$
        \EndIf
        \State $\alpha^*_{sk} \gets \alpha_{sk}$,
        \State $ModelAssign[IDK] \gets k+1$,
        \State $cost += \sum_{IDK} \mathcal{C}[m_{sk+1}] - \mathcal{C}[m_{k}]$
    \EndIf
\EndFor
\State  $\bm{\alpha}^*_s \gets \bm{\alpha}^*_s + [\alpha^*_{sk}]$
\EndFor
\State \Return $\bm{\alpha}^*_s$
\EndProcedure
\end{algorithmic}
\end{algorithm}

\begin{algorithm}[]
 \caption{\textit{Soft-gating} Algorithm for In-Stage $IDK$ Cascade}
\begin{algorithmic}[1]
\Statex \textbf{Input}
\Statex \hspace*{\algorithmicindent} $\mathcal{D}^s$: Training data containing $N^s$ samples in stage-$s$
\Statex \hspace*{\algorithmicindent} $\mathcal{M}^s$: Sorted list of the models trained for stage-$s$
\Statex \hspace*{\algorithmicindent} $\bm{f}_{\mathcal{M}^s}$: A multi-head DKD model for distilling all the model's confidence at stage-$s$
\Statex \hspace*{\algorithmicindent} $\mathcal{C}$: Dictionary of models'  spatio-temporal costs
\Statex \hspace*{\algorithmicindent} $c_s$: User-defined budget of spatio-temporal cost for stage-$s$
\Statex \hspace*{\algorithmicindent} $q$: Confidence function
\Statex \hspace*{\algorithmicindent} $\lambda$: Controller for the spatio-temporal cost budget
\Statex \hspace*{\algorithmicindent} $\mu$: Controller for L1-norm sparsity regularization
\Statex \hspace*{\algorithmicindent} $nEpochs$: Number of training epochs
\Statex \textbf{Output}
\Statex \hspace*{\algorithmicindent} $\bm{a}^*_s,\bm{b}^*_s$: the optimal soft-gating IDK coefficient for stage-s
\Procedure{SoftGating}{$\mathcal{D}^s$, $\mathcal{M}^s$, $\bm{f}_{\mathcal{M}^s}$, $c_s$, $\mathcal{C}$, $q$}
\State $lr \gets 1e-1$, $e \gets 0$, $\bm{a}_s \gets 1$, $\bm{b}_s \gets 0.5$
\While {e < nEpochs}
\State $\hat{q}_{sj}(\bm{x}_t) \gets q\big(\bm{f}(\bm{x}_t; m_{sk})[1]/\sum \bm{f}(\bm{x}_t; m_{sk})\big)$ \Comment{DKD confidence distillation}
\State $\mathcal{L}^s_{\text{sparse}} \gets \sum_{i,t,k} \mid\mid G^{IDK}(\hat{q}_{sj}(\bm{x}_t) )\mid\mid_1$
\State $\mathcal{L}_{\bm{a}_s, \bm{b}_s} \gets \mathcal{L}^s_{\text{nll}} + \lambda \mathcal{L}^{s}_{\text{cost}} + \mu \mathcal{L}^s_{\text{sparse}}$
\State Optimize $\mathcal{L}_{\bm{a}_s, \bm{b}_s}$ using SGD
\State Reduce $lr$ by factor $0.5$ once learning stagnates.
\State $e \gets e + 1$
\EndWhile
\State \Return $\bm{a}^*_s,\bm{b}^*_s$
\EndProcedure
\end{algorithmic}
\end{algorithm}

\begin{algorithm}[]
 \caption{End-to-End Training algorithm for \system}
\begin{algorithmic}[1]
\Statex \textbf{Input}
\Statex \hspace*{\algorithmicindent} $\mathcal{D}$: Full training data containing $N$ instances
\Statex \hspace*{\algorithmicindent} $\mathcal{M}$: Full model zoo
\Statex \hspace*{\algorithmicindent} $\mathcal{C}$: Dictionary of models' spatio-temporal costs
\Statex \hspace*{\algorithmicindent} $q$: Confidence criterion
\Statex \textbf{Output}
\Statex \hspace*{\algorithmicindent} $\bm{\theta}^*$: the optimal ICK$_1$ gate parameters
\Statex \hspace*{\algorithmicindent} $\bm{\alpha}^*$ (or $\bm{a}^*, \bm{b}^*$): the optimal IDK gate parameters
\Procedure{End-to-EndTraining}{$\mathcal{D}$, $\mathcal{M}$}
\State Pre-allocate costs $c_s$ for each stage s.
\State \textbf{Step 1:} Learn in-stage IDK gate parameters.
\For{each stage s}
    \State $\bm{\alpha}^* \gets $ HardGating($\mathcal{D}^s, \mathcal{M}^s, c_s, C, q$)
    \State or, $\bm{a}^*, \bm{b}^* \gets $ SoftGating($\mathcal{D}^s, \mathcal{M}^s, c_s, C, q$)
\EndFor
\\\hrulefill
\State \textbf{Step 2:} Learn ICK$_1$ gate parameters.
\For{each model $m_{sk}$}
\State $\theta^*_{sk} \gets $ Grid Search for minimizing $\sqrt{\epsilon^2_{\text{ICK}_1} + \epsilon^2_{\text{ICK}_0}}$
\EndFor
\State \Return $\bm{\alpha}^*$ (or $\bm{a}^*, \bm{b}^*$),  $\bm{\theta}^*$
\EndProcedure
\end{algorithmic}
\end{algorithm}

\subsection{Data preparation}
\label{sec:data}
By following the definition of Sepsis-3 \cite{singer2016third}, we identify the sepsis onset to be the time when an increase in the Sequential Organ Failure Assessment (SOFA) score of $2$ points or more occurs in response to infections.
We use the \textit{Sepsis-3} toolkit\footnote{\url{https://doi.org/10.5281/zenodo.1256723}} to obtain the suspected infection time in patients,
and following the process in \cite{seymour2016assessment} to finally label the onset of sepsis.
We result at a total number of $20,009$ sepsis patients out of the $52,902$ adult patients from MIMIC-III database. We exclude those patients who stay in ICUs less than $6$ hours and also exclude those patients who developed sepsis within the first $6$ hours after ICU admission. This reduces our cohort to a total of $34,475$ ICU patient, and only $2,370 (6.8\%)$ out of them are labeled as sepsis (because $88.1\%$ of sepsis onsets happened within the first 6 hours after ICU admission and are excluded from our study cohort). Then according to \cite{singer2016third}, we identify the onset of septic shock as when a vasopressor is  required to maintain a mean arterial pressure (MAP) $\geq 65$ mm Hg and serum lactate level $> 2$ mmol/L ($>18$ mg/dL). We result at $229 (9.7\%)$ septic shock patients out of the $2,370$ sepsis patients.

For feature generation, we extract 8 patient static characteristics including age, gender, race, height, weight, sepsis onset hour since ICU admission, whether diagnosed diabetes or on a ventilator at ICU admission. Then we extract the dynamic features by obtaining the 8 vital signs, 16 lab measurements, 6 vassopressors, continuous replacement therapies (CRRT), ventilation, 2 intravenous fluids in fluid resuscitation, and 5 additional measurements that are recommended for monitoring during sepsis management.
The 8 vital signs include heart rate, systolic blood pressure, diastolic blood pressure, mean blood pressure, respiration rate, temperature, SpO2 and glucose. The 16 lab measurements include Anion gap, Albumin, Bands, Bicarbonate, Bilirubin, Creatinine, Chloride, Glucose, Hematocrit, Hemoglobin, Lactate, Platelet, Potassium, PTT, INR, PT, Sodium, BUN and WBC. The 6 vasopressors include dobutamine, dopamine, epinephrine, norepinephrine, phenylephrine, and vasopressin. The 2 fluids include Crystalloids and Colloids that are recommended in the early management of sepsis \cite{rhodes2017surviving}, and particularly
fluid resuscitation of bolus $\geq 500$ mL is one of the most common treatment for managing septic shock.
The 5 additional measurements include whether a vasopressor is needed to maintain a mean arterial pressure (MAP) $\geq$ 65 mm Hg, serum lactate level $>$ 2 mmol/L, urine output $\geq$ 5 ml/kg/hr, venous oxygen saturation (SvO2) $\geq 70\%$ and central venous pressure (CVP) of $8 - 12$ mmHg.
We fill missing values like lab measurements using the last measured value; we clamp real-valued features in between their $0.05$-quantile and $0.95$-quantile values respectively and normalize the features using min-max normalization. 

\begin{figure}
  \centering
  \includegraphics[width=0.5\linewidth]{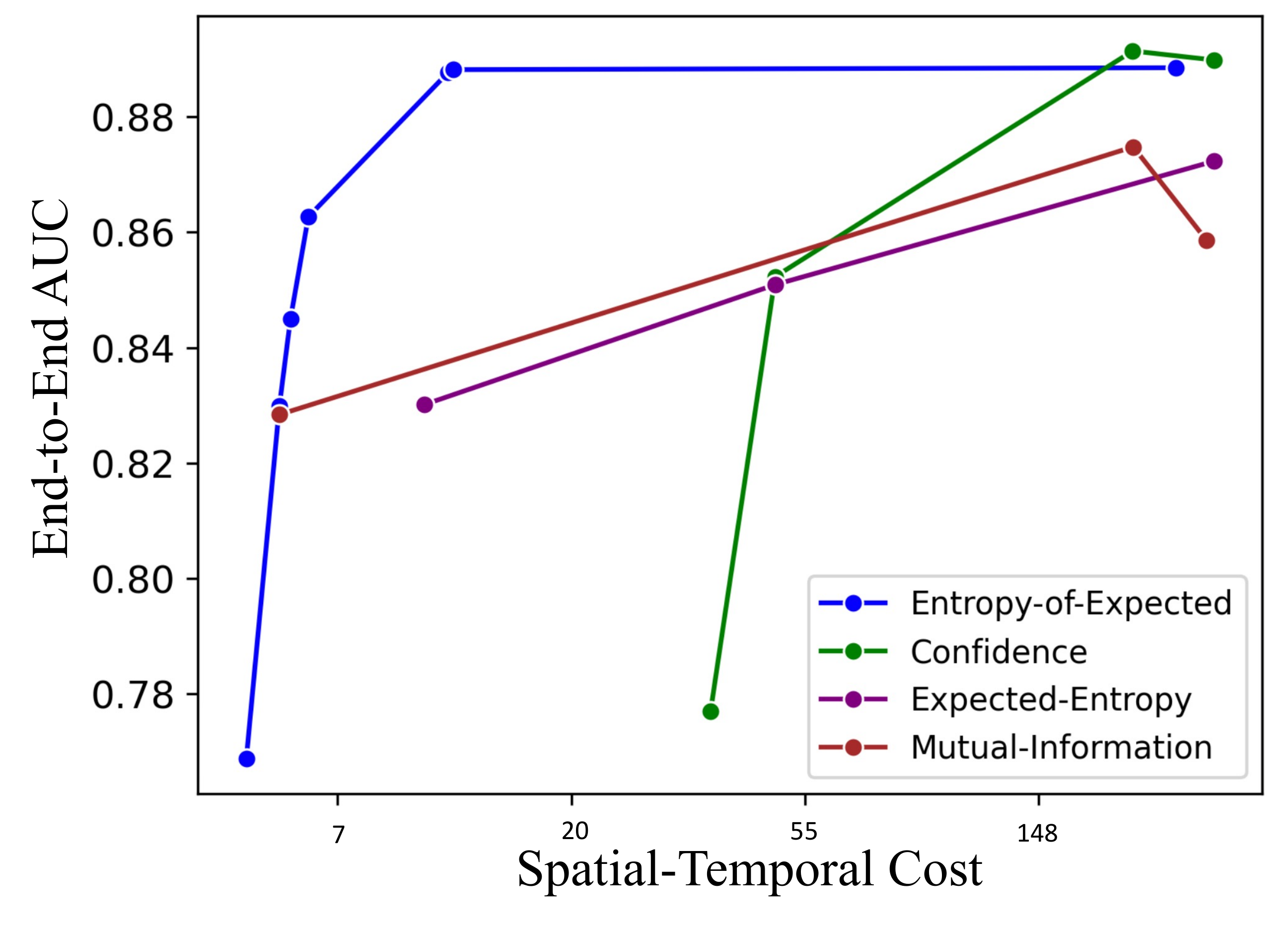}
  \vspace{-1em}
  \caption{\small Confidence measure selection in Soft-Gating}
  \vspace{-1em}
  \label{fig:m_search}
\end{figure}
\begin{figure}
  \centering
  \includegraphics[width=0.7\linewidth]{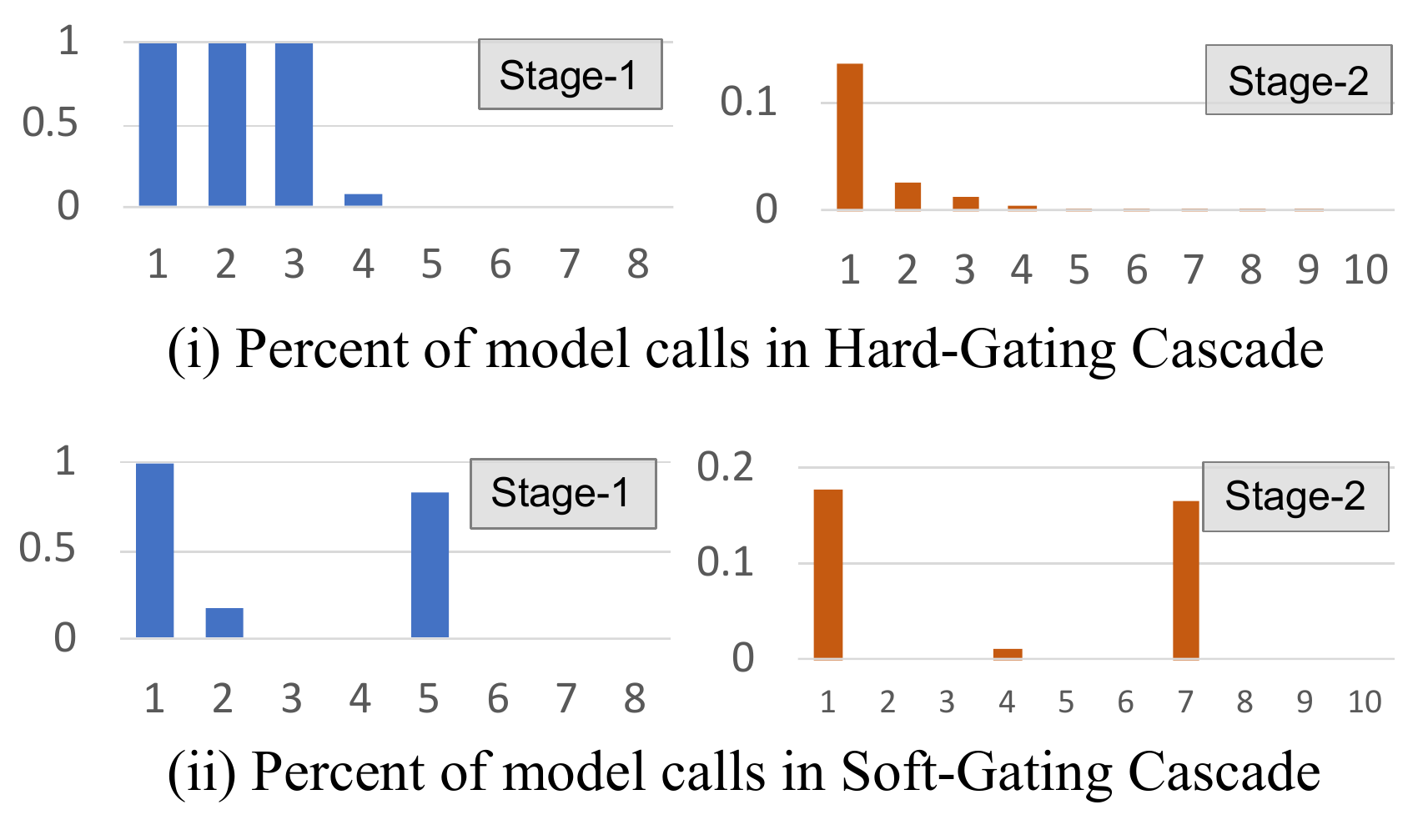}
  \caption{\small Transitions in model calls: both cascades always call the first model per each stage for an entrance and transition to next models (IDK) or next stage (ICK).}
  \vspace{-1em}
  \label{fig:calls}
\end{figure}
For training sepsis prediction models, we take the full training cohort but discard the data after the first sepsis onset in sepsis patients, then we label the data per hour, and label the current sepsis outcome as 1 if the true sepsis is going to happen in the next 12 hours (designed for early prediction on sepsis). For training shock prediction models, we take the sepsis sub training cohort and discard the data before sepsis onset. We also discard the data after septic shock onset in shock patients. Then we label in the same way as for sepsis, i.e. label the current shock outcome as 1 if the true shock will take place in the next 12 hrs. For those non sepsis patients, we discard the first 12 hrs data after ICU admission to reduce data noises and randomly sample a sequence length between 12 hrs up to 7 days per each non sepsis patients. More details of data prepossessing are provided in the attached code.

\subsection{Model Zoo}
\label{sec:zoo}
Computational cost was measured in $ms$ as the total running time of feeding all the test data (with batch size of 256) calling each individual models on a single \texttt{GeForce RTX 2080Ti} divided by the total number of calls. Then we multiply the cost by 10 as the GPU is approximately 10X hardware cost comparing to a CPU. Future work can extend the model zoo to include CPU models or running all the models on CPUs based on resource specifications. 
Table \ref{tab:sepsis_models} and Table \ref{tab:shock_models} respectively show the model prediction AUC scores on the validation set for the sepsis and septic shock stages. 

In addition, we also fit small DKD surrogate models for distilling the predictive probabilities and confidence of the models in the zoo. The DKD model is a 4-layer MLP taking the embedding vectors from the first model in each stage, so it obtains similar AUC scores on the validation set comparing to the early models in the zoo but much lower scores comparing to the later heavier models. But the mean absolute errors (MAE) of the DKD model on estimating confidence measures of the original models are consistently small, which is beneficial for our soft-gating algorithm that requires only confidence estimation instead of predictive probabilities.  

\begin{table*}[]
\resizebox{\textwidth}{!}{%
\begin{tabular}{l|llll|lllll}
\toprule
 & \multicolumn{4}{c|}{Model Zoo} & \multicolumn{5}{c}{Dirichlet Knowledge Distillation (DKD)} \\
 \midrule
Model & AUC & Computational & Data & Total & AUC & MAE & MAE & MAE & MAE\\
& & Cost & Modality & Norm. Cost  & & Confidence & Entropy of Exp. & Entropy & MI\\
\midrule
vitals\_1hr.h100.nlayer1 & 74.5\% & 5 & 1 & 0.10 & 74.4\% & 0.07 & 0.13 & 0.08 & 0.05 \\
vitals\_6hr.h100.nlayer1 & 78.2\% & 7 & 1 & 0.11 & 74.7\% & 0.06 & 0.11 & 0.07 & 0.05 \\
vitals\_6hr.h100.nlayer3 & 79.7\% & 172 & 1 & 0.54 & 74.8\% & 0.08 & 0.14 & 0.09 & 0.06 \\
vitals\_6hr.h300.nlayer2 & 81.1\% & 173 & 1 & 0.54 & 75.0\% & 0.08 & 0.14 & 0.09 & 0.06 \\
vitals\_12hr.h200.nlayer4 & 82.3\% & 175 & 1 & 0.55 & 70.5\% & 0.09 & 0.16 & 0.10 & 0.10 \\
vitals\_labs\_1hr.h100.nlayer1 & 76.8\% & 5 & 2 & 0.20 & 74.0\% & 0.06 & 0.11 & 0.07 & 0.04 \\
vitals\_labs\_6hr.h100.nlayer1 & 81.8\% & 86 & 2 & 0.41 & 74.0\% & 0.06 & 0.12 & 0.08 & 0.04 \\
vitals\_labs\_6hr.h100.nlayer2 & 82.6\% & 257 & 2 & 0.86 & 73.5\% & 0.06 & 0.12 & 0.08 & 0.05 \\
vitals\_labs\_6hr.h100.nlayer3 & 82.5\% & 258 & 2 & 0.86 & 74.2\% & 0.08 & 0.14 & 0.09 & 0.05 \\
vitals\_labs\_csu\_1hr.h100.nlayer1 & 78.3\% & 5 & 3 & 0.30 & 73.8\% & 0.07 & 0.13 & 0.09 & 0.05 \\
vitals\_labs\_csu\_6hr.h100.nlayer1 & 81.6\% & 90 & 3 & 0.52 & 73.4\% & 0.08 & 0.14 & 0.10 & 0.05 \\
vitals\_labs\_csu\_6hr.h400.nlayer1 & 81.6\% & 258 & 3 & 0.96 & 73.5\% & 0.05 & 0.11 & 0.07 & 0.05 \\
vitals\_labs\_csu\_6hr.h400.nlayer3 & 83.5\% & 259 & 3 & 0.97 & 73.7\% & 0.07 & 0.13 & 0.08 & 0.06 \\
vitals\_labs\_csu\_6hr.h300.nlayer3 & 82.2\% & 264 & 3 & 0.98 & 73.5\% & 0.07 & 0.14 & 0.08 & 0.07 \\
vitals\_labs\_csu\_6hr.h100.nlayer2 & 81.8\% & 272 & 3 & 1.00 & 73.2\% & 0.09 & 0.15 & 0.10 & 0.06 \\
vitals\_labs\_csu\_12hr.h300.nlayer4 & 85.1\% & 268 & 3 & 0.99 & 72.6\% & 0.09 & 0.16 & 0.10 & 0.07\\
\bottomrule
\end{tabular}}
\caption{Sepsis-Stage model zoo}
\label{tab:sepsis_models}
\end{table*}

\begin{table*}[]
\resizebox{\textwidth}{!}{%
\begin{tabular}{l|llll|lllll}
\toprule
 & \multicolumn{4}{c|}{Model Zoo} & \multicolumn{5}{c}{Dirichlet Knowledge Distillation (DKD)} \\
 \midrule
Model & AUC & Computational & Data & Total & AUC & MAE & MAE & MAE & MAE\\
& & Cost & Modality & Norm. Cost  & & Confidence & Entropy of Exp. & Entropy & MI\\
\midrule
vitals\_1hr.h100.nlayer1 & 87.0\% & 5 & 1 & 0.23 & 87.1\% & 0.05 & 0.07 & 0.04 & 0.03 \\
vitals\_6hr.h100.nlayer1 & 88.6\% & 7 & 1 & 0.23 & 86.7\% & 0.04 & 0.08 & 0.06 & 0.03 \\
vitals\_6hr.h100.nlayer3 & 88.4\% & 172 & 1 & 0.28 & 86.9\% & 0.04 & 0.08 & 0.06 & 0.03 \\
vitals\_6hr.h300.nlayer2 & 86.8\% & 173 & 1 & 0.28 & 86.5\% & 0.04 & 0.07 & 0.06 & 0.03 \\
vitals\_12hr.h300.nlayer2 & 88.6\% & 174 & 1 & 0.29 & 86.0\% & 0.04 & 0.09 & 0.07 & 0.03 \\
vitals\_12hr.h200.nlayer4 & 88.6\% & 175 & 1 & 0.29 & 85.3\% & 0.04 & 0.09 & 0.06 & 0.04 \\
vitals\_12hr.h300.nlayer3 & 85.1\% & 177 & 1 & 0.29 & 85.3\% & 0.05 & 0.11 & 0.08 & 0.04 \\
vitals\_12hr.h400.nlayer3 & 89.5\% & 189 & 1 & 0.29 & 85.6\% & 0.03 & 0.07 & 0.05 & 0.02 \\
vitals\_labs\_1hr.h100.nlayer1 & 89.0\% & 5 & 2 & 0.45 & 85.4\% & 0.03 & 0.06 & 0.04 & 0.03 \\
vitals\_labs\_6hr.h100.nlayer1 & 89.8\% & 86 & 2 & 0.48 & 86.1\% & 0.04 & 0.08 & 0.05 & 0.04 \\
vitals\_labs\_6hr.h100.nlayer2 & 89.9\% & 257 & 2 & 0.54 & 85.4\% & 0.03 & 0.06 & 0.04 & 0.04 \\
vitals\_labs\_6hr.h300.nlayer1 & 87.7\% & 258 & 2 & 0.54 & 84.0\% & 0.03 & 0.07 & 0.04 & 0.04 \\
vitals\_labs\_6hr.h200.nlayer2 & 89.8\% & 263 & 2 & 0.54 & 87.4\% & 0.04 & 0.09 & 0.06 & 0.04 \\
vitals\_labs\_12hr.h300.nlayer4 & 93.5\% & 262 & 2 & 0.54 & 82.9\% & 0.01 & 0.03 & 0.02 & 0.01 \\
vitals\_labs\_12hr.h200.nlayer4 & 90.7\% & 270 & 2 & 0.54 & 89.1\% & 0.01 & 0.04 & 0.02 & 0.02 \\
vitals\_labs\_csu\_1hr.h100.nlayer1 & 90.8\% & 5 & 3 & 0.68 & 86.2\% & 0.04 & 0.09 & 0.06 & 0.04 \\
vitals\_labs\_csu\_6hr.h100.nlayer1 & 91.9\% & 90 & 3 & 0.71 & 86.3\% & 0.04 & 0.10 & 0.06 & 0.05 \\
vitals\_labs\_csu\_1hr.h100.nlayer4 & 90.2\% & 172 & 3 & 0.73 & 86.7\% & 0.02 & 0.05 & 0.04 & 0.02 \\
vitals\_labs\_csu\_6hr.h200.nlayer2 & 88.9\% & 258 & 3 & 0.76 & 86.3\% & 0.02 & 0.06 & 0.04 & 0.03 \\
vitals\_labs\_csu\_6hr.h300.nlayer3 & 88.7\% & 264 & 3 & 0.77 & 88.0\% & 0.01 & 0.02 & 0.01 & 0.01 \\
vitals\_labs\_csu\_12hr.h200.nlayer3 & 92.1\% & 287 & 3 & 0.78 & 86.2\% & 0.02 & 0.05 & 0.03 & 0.02 \\
vitals\_labs\_csu\_med\_1hr.h100.nlayer1 & 91.5\% & 5 & 4 & 0.90 & 85.5\% & 0.03 & 0.06 & 0.03 & 0.05 \\
vitals\_labs\_csu\_med\_6hr.h100.nlayer1 & 91.6\% & 86 & 4 & 0.93 & 87.3\% & 0.03 & 0.07 & 0.04 & 0.04 \\
vitals\_labs\_csu\_med\_6hr.h400.nlayer1 & 91.4\% & 257 & 4 & 0.99 & 87.1\% & 0.03 & 0.08 & 0.05 & 0.04 \\
vitals\_labs\_csu\_med\_6hr.h300.nlayer3 & 90.4\% & 259 & 4 & 0.99 & 86.4\% & 0.03 & 0.07 & 0.05 & 0.03 \\
vitals\_labs\_csu\_med\_12hr.h100.nlayer2 & 93.4\% & 262 & 4 & 0.99 & 83.3\% & 0.00 & 0.01 & 0.01 & 0.01 \\
vitals\_labs\_csu\_med\_12hr.h400.nlayer4 & 93.4\% & 269 & 4 & 0.99 & 84.7\% & 0.02 & 0.06 & 0.02 & 0.04\\
\bottomrule
\end{tabular}}
\caption{Septic Shock-Stage model zoo}
\label{tab:shock_models}
\end{table*}

\subsection{Model Utilization in \system.}
\label{sec:others}
 We analyze model utilization frequency (the proportion of how many times a model was invoked) in our test cohort and compare model frequencies for hard and soft gating in Figure \ref{fig:calls} (models are grouped into 8 groups for Stage 1 and 10 groups for Stage 2): soft gating can skip invocations of many models and directly select the more confident models for faster transitioning to shock stage.






\noindent where $\psi$ is the \textit{digamma} function defined as the logarithmic derivative of the gamma function, $\alpha_0$ and $\alpha_1$ are the concentration parameters estimated by the DKD models. More definitions are in \citet{malinin2018predictive}. We show ``Entropy of expected'' exhibits the best AUC-Cost trade-off path in Figure \ref{fig:m_search}. 

\subsection{Qualitative Evaluation}
\label{sec:exp:demo}
\begin{figure}
   \centering
    \includegraphics[width =.6\linewidth]{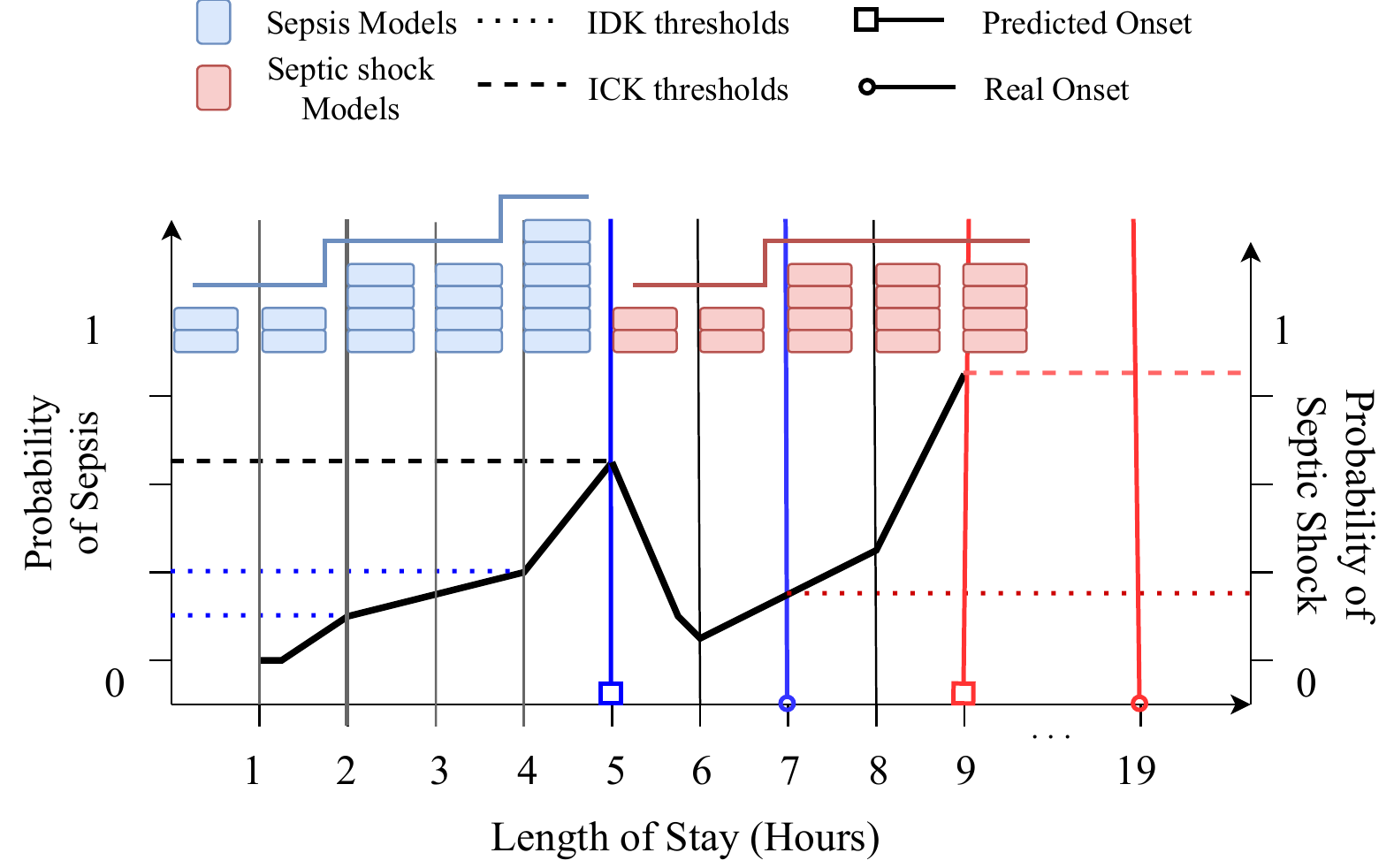}
    \vspace{-0.5em}
    \caption{\small A timeline shown for an example shock patient. The y-axis represents probabilities of sepsis (left) and septic shock (right). The x-axis represents patient's length of stay (hours). This figure illustrates how different models are selected based on patient's critical health condition and timely septic shock prediction is made in cost-efficient manner.}
    \label{fig:demo}
\end{figure}

We walk through an example shock patient's length of stay in ICU from the test set, and deploy the proposed multi-stage prediction pipeline on it. \system starts the prediction of sepsis with a cheaper model as seen in Figure \ref{fig:demo}. At t=2, the model's prediction probability reaches the $IDK$ threshold which signifies model's uncertainty in sepsis prediction. Hence, the \system switches to a costlier and more accurate model (a similar trend is observed at t=4,7). At t=5, \system predicts sepsis onset as the probability of sepsis prediction reaches $ICK$ threshold. Note, once sepsis onset is predicted by the cascade, it switches to a cheaper model which predicts septic shock (Stage-2). Due to early switching, \system can detect septic shock significantly earlier. In Stage-2, \system transitions to costlier model once the cheaper model becomes uncertain. Lastly, it predicts septic shock once the probability of septic shock detection reaches the $ICK_1$ threshold. 

\begin{figure}
    \centering
  \includegraphics[width=0.6\linewidth]{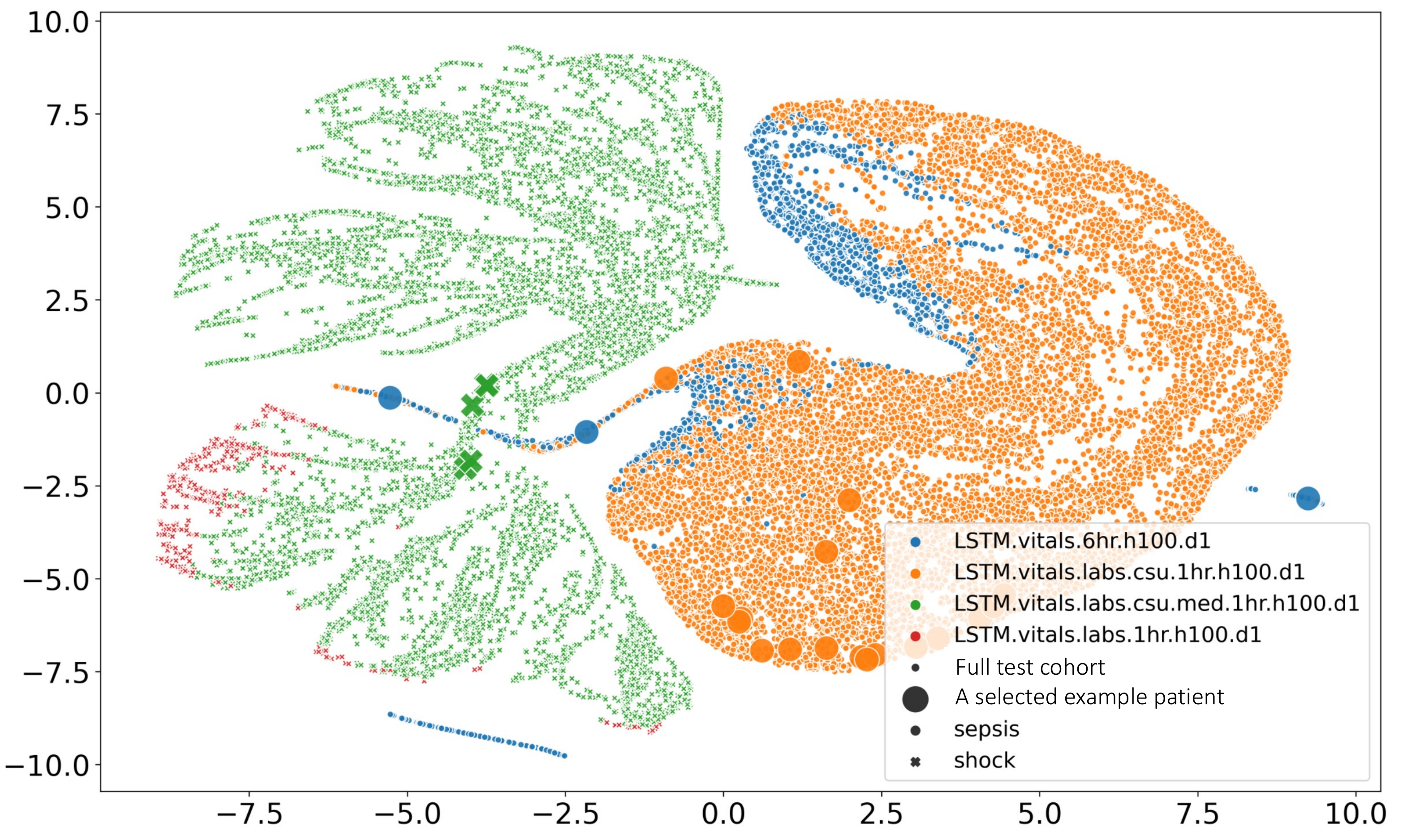}
  \vspace{-0.5em}
  \caption{\small Dynamic model allocations in the \system: the example shock patient (in large-sized marker) transitioned from cheaper model in sepsis (dot) stage to costlier model in shock stage (cross).}
  \label{fig:tsne}
  \vspace{-0.5em}
\end{figure}

Additionally, we randomly slice $35$k time-steps from the sequential data in the test set and visualize them in a TSNE \cite{van2008visualizing} plot in Figure \ref{fig:tsne} based on their embedding vectors generated from the LSTMs in the model zoo. Different colors show the different model allocations for the subsampled test data points, sepsis (dot) and shock (cross) stages are clearly separated in to the left and right regions of the 2-D transformation space. We highlight the picked shock patient (with significantly large markers) showing its dynamic model allocations and stage transitions within \system.

\end{document}